\documentclass{bmvc2k}

\usepackage{graphicx}
\usepackage{tikz}
\usepackage{comment}
\usepackage{amsmath,amssymb} 
\usepackage{color}
\usepackage{verbatim}
\usepackage{subcaption}
\usepackage{booktabs}
\usepackage{enumitem}
\usepackage{algorithm2e}
\usepackage{lipsum}
\usepackage{xfrac}
\usepackage{soul}

\usepackage[xindy]{glossaries}

\newcommand\blfootnote[1]{%
  \begingroup
  \renewcommand\thefootnote{}\footnote{#1}%
  \addtocounter{footnote}{-1}%
  \endgroup
}

\title{Can I see an Example? Active Learning the Long Tail of Attributes and Relations}

\addauthor{Tyler~L.~Hayes}{tlh6792@rit.edu}{1$^\ast$}
\addauthor{Maximilian~Nickel}{maxn@fb.com}{2}
\addauthor{Christopher~Kanan}{ckanan@cs.rochester.edu}{1,3}
\addauthor{Ludovic~Denoyer}{ludovic.den@gmail.com}{2$^\dagger$}
\addauthor{Arthur~Szlam}{aszlam@fb.com}{2}

\addinstitution{
 Rochester Institute of Technology\\
 Rochester, NY, USA
}
\addinstitution{
 Facebook AI Research\\
 New York, NY, USA
}
\addinstitution{
 University of Rochester\\
 Rochester, NY, USA
}

\runninghead{Hayes, Nickel, Kanan, Denoyer, Szlam}{Can I see an Example?}

\usepackage{ifthen}
\newboolean{ack}
\newboolean{combined}
\setboolean{ack}{true} 
\setboolean{combined}{true}

\newcommand{\beginsupplement}{%
        \setcounter{table}{0}
        \renewcommand{\thetable}{S\arabic{table}}%
        
        \setcounter{figure}{0}
        \renewcommand{\thefigure}{S\arabic{figure}}%
    
        \setcounter{equation}{0}
        \renewcommand{\theequation}{S\arabic{equation}}%
        
        \setcounter{section}{0}
        \renewcommand{\thesection}{S\arabic{section}}%
        
        \setcounter{page}{1}
        \renewcommand{\thepage}{S\arabic{page}}%
}

\newglossaryentry{SPO}
{
name={$\left(s, p, o\right)$},
description={$\left(s, p, o\right)$}
}
\newglossaryentry{SPA}
{
name={$\left(s, p, a\right)$},
description={$\left(s, p, a\right)$}
}
\newglossaryentry{SPAA}
{
name={$\left(s, p, a?\right)$},
description={$\left(s, p, a?\right)$}
}
\newglossaryentry{SPAP}
{
name={$\left(s, ?, a\right)$},
description={$\left(s, ?, a\right)$}
}
\newglossaryentry{SPAS}
{
name={$\left(?, p, a\right)$},
description={$\left(?, p, a\right)$}
}
\newglossaryentry{SPOO}
{
name={$\left(s, p, o?\right)$},
description={$\left(s, p, o?\right)$}
}
\newglossaryentry{SPOP}
{
name={$\left(s, ?, o\right)$},
description={$\left(s, ?, o\right)$}
}
\newglossaryentry{SPOS}
{
name={$\left(?, p, o\right)$},
description={$\left(?, p, o\right)$}
}

\begin{document}

\maketitle
\sloppy
\blfootnote{$^\ast$ Work done while interning at Facebook AI Research}
\blfootnote{$^\dagger$ Now at Ubisoft}

\begin{abstract}
There has been significant progress in creating machine learning models that identify objects in scenes along with their associated attributes and relationships; however, there is a large gap between the best models and human capabilities. One of the major reasons for this gap is the difficulty in collecting sufficient amounts of annotated relations and attributes for training these systems. While some attributes and relations are abundant, the distribution in the natural world and existing datasets is long tailed. In this paper, we address this problem by introducing a novel incremental active learning framework that asks for attributes and relations in visual scenes. While conventional active learning methods ask for labels of specific examples, we flip this framing to allow agents to ask for examples from specific categories. Using this framing, we introduce an active sampling method that asks for examples from the tail of the data distribution and show that it outperforms classical active learning methods on Visual Genome.
\end{abstract}

\section{Introduction}

In active learning~\cite{lin2017active,settles2009active,wang2016cost,wei2015submodularity,yoo2019learning}, a learning agent is provided with unlabeled samples, from which it selects those it considers critical to improving its performance to be labeled by a teacher (oracle). By selecting the most informative samples to be labeled, it is hoped that the agent can achieve better performance with less data.  However, in practice, the extra complexity of active learning approaches often outweigh their benefits.    In our view, the power of active learning is tempered by the framing of the classical machine learning problems where it is often applied.  Namely, a classification problem with few classes, with loss measured in average classification accuracy on the test distribution, and a repertoire of questions of the form ``what is the label of example $i$?''.  This is in contrast to the ``active learning'' of humans, who have access to extremely rich supervision beyond asking for the label of an example.

\begin{figure}[t]
\begin{center}
  \includegraphics[width=0.5\linewidth]{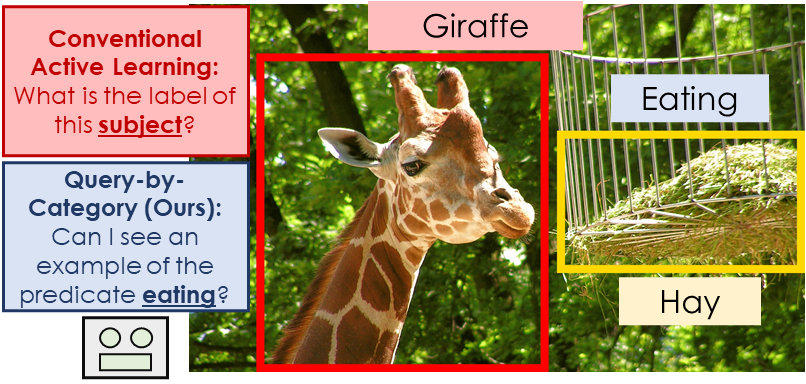}
\end{center}
\vspace{0.1in}
  \caption{
In classical active learning, an agent selects \emph{examples} that it is most uncertain about to be labeled. Conversely, we allow agents to ask for examples at the \emph{category} level instead of the example level. We train agents to predict attributes and relations in scenes. Our framing allows an agent to ask for an example of a rare category ``eating'' and an oracle provides an example of this predicate. \label{fig:main}
}
\end{figure}

In this work, we study the problem of learning to predict objects, attributes, and relations in visual scenes. The set of possible attributes and relations is long-tailed (that is, there are many attributes and relations, with many categories containing very few samples), and its structure suggests more sophisticated questions a learning agent might ask. For example, an agent might ask for an example from a particular \emph{label} it is uncertain about, instead of a particular sample it is uncertain about. By asking for class labels instead of instances, an agent can explicitly sample from more underrepresented classes, which is useful for long-tailed data. We show that several classical active learning methods do not perform significantly better than random sampling; and in particular fail to explore the tail of the distribution. By simply switching the framing to allow the agent to ask for an example for an attribute or relation (rather than ask for the true attribute or relation for object instances in an image), we can significantly improve results on the tail of the distribution, without sacrificing accuracy on the natural distribution (see Fig.~\ref{fig:main}). This is critical as visual scenes are typically long-tailed and we desire methods capable of performing well on the tail of the distribution.

\textbf{We make the following major contributions:} We introduce a novel incremental active learning framework, coined ``Query-by-Category'' (QBCat), that allows agents to ask for labels at the \emph{category} level instead of the \emph{example} level. To demonstrate the effectiveness of our framework, we study active learning on a new domain: training agents to predict objects, attributes, and relations in visual scenes. Since the distribution of attributes and relations is naturally long-tailed, we introduce two methods to enable incremental long-tailed learning including class re-balancing and bias correction. Finally, we use our Query-by-Category framework to introduce a new active sampling approach that samples efficiently from the tail of the attribute and relation class distributions. We experimentally validate the effectiveness of our setup, which outperforms classical active learning methods on Visual Genome~\cite{krishnavisualgenome}.

\textbf{Motivation for Query-by-Category Learning.} In conventional active learning, the agent computes uncertainty scores over unlabeled examples to select examples to be labeled. While straightforward, this approach has drawbacks: a representative set of unlabeled examples must be collected, the learner must compute uncertainty scores for the examples, and the learner must have robust uncertainty estimates to select examples. These tasks can be time consuming for both annotators and the learner. Moreover, the gamut of queries the learner can ask for is limited, i.e., the model simply asks for the label of an unlabeled example.

We argue that it could be beneficial to allow a model to ask directly for examples from specific classes. This eliminates the need for model inference on unlabeled data. Further, by asking for classes directly, the agent will likely see a wider variety of classes, which is critical in long-tailed settings. Just as in conventional active learning, the primary focus of our framework is on learning efficiency rather than the speed in which an oracle can provide annotations. Moreover, in a practical implementation, an annotator could be augmented with a search engine or lots of thumbnail images to quickly find images of a specific class. Alternatively, the learner could be provided with a search engine or database to gather labeled data for itself. We leave these implementations for future work; here we are first interested in determining if our framing yields performance benefits over conventional active learning.

\section{Methods}
\label{sec:methods}

We operate on \emph{(subject, predicate, object)} triples in images.  The subject will be a bounding box marking a region of pixels (e.g., an object) in an image. The object might be another bounding box if the predicate is a spatial relationship, like ``to the left of''; or a string like ``blue'' (which could correspond to a predicate like ``has color''). If the object is a string, we use the notation \emph{(subject, predicate, attribute)}. We do not consider higher arity relations or relations that cross multiple images. The learner will be tasked with predicting the missing element in an incomplete triple. We denote the triple with two known elements as the \emph{question} and the correct missing element as the \emph{target}. We use $s$, $o$, $p$, and $a$ to denote subjects, objects, predicates, and attributes respectively. This yields six unique question types: \gls{SPOS} and \gls{SPAS} where the target is a subject, \gls{SPOP}, and \gls{SPAP} where the target is a predicate, \gls{SPOO} where the target is an object, and \gls{SPAA} where the target is an attribute.

Rather than assuming the labeled data is fixed at the start of learning, we perform incremental active learning, i.e., the agent asks for a fixed number of annotations over a sequence of increments. However, we will see that classical active learning techniques do not significantly improve on random selection, in part due to the long tail of attributes and relations. Our main result is that asking for examples of a triple (described in Sec.~\ref{sec:tail-sampling}) is more effective at exploring the tail of possibilities than asking for triple completions. However, this active sampling strategy biases the model to the tail and reduces performance on the natural distribution. We will see that combined with proper re-biasing techniques, we get the best of both worlds: improved tail performance without sacrificing accuracy on the natural distribution. 

\subsection{Incremental Training Procedure}
\label{sec:train-protocol}

Let $\mathcal{D}_{L}$ and $\mathcal{D}_{U}$ denote labeled and unlabeled subsets of the dataset $\mathcal{D} = \mathcal{D}_{L} \cup \mathcal{D}_{U}$, respectively, where $\mathcal{D}$ is a dataset of triples.  The ``unlabeled'' triples correspond to the ``questions'' from above; whereas ``labeled'' examples are the complete triple (question and target together).  For the rest of the paper, we will use ``natural distribution'' to refer to either the base distribution on $\mathcal{D}$ or to targets in $\mathcal{D}$; and the ``head'' to refer to the elements in $\mathcal{D}$ with the most common targets (or to those targets themselves), and the ``tail'' to refer to other targets.

In the real world, an agent would receive a stream of inputs for which it could incrementally ask questions to improve its performance. Motivated by this, we perform incremental active learning as follows. We first pre-train the model on $\mathcal{D}_{L}$ drawn from the head of the training data. We wish to simulate the setting where we first collect some seed data (which, following the natural distribution, would mostly come from the head), and then the agent actively learns starting from a model pre-trained on the seed data. After pre-training, we initialize a replay buffer with all pre-training data. Then, an active sampling strategy (described in Sec.~\ref{sec:active-sampling}) selects $B$ samples from the unlabeled dataset ($\mathcal{D}_{U}$) to be labeled by an oracle.

After an oracle labels the $B$ samples, training happens in two stages: cross-validation and full training. Cross-validation determines the number of epochs for full training to prevent over-fitting on new samples (see Sec.~\ref{sec:cross-valid}). After cross-validation, we reset model parameters to the beginning of the increment, retrain with the validated hyper-parameters, and optionally re-bias to the natural distribution (see Sec.~\ref{sec:methods-imbalance}). We then evaluate the model (see Sec.~\ref{sec:experimental-protocol}) and put all newly labeled examples in the replay buffer and repeat. The process of adding new data, cross-validation, training, and optional re-biasing is called an ``increment.''

\subsection{Methods to Handle Class Imbalance} 
\label{sec:methods-imbalance}

Because of the long-tailed data distribution, naive training on increments leads to over-fitting on the head classes and hinders learning after the first increment (see Sec.~\ref{sec:additional-study-standard-mb}).
To address this, we re-balance mini-batches during increments such that there are equal amounts of old (frequently represented) data and new (possibly less frequently represented) data (see Sec.~\ref{sec:re-balance}). This allows learning past the first increment; but results in the model learning a distribution different from the natural distribution. To remedy this, we perform post-hoc bias correction~\cite{hong2021disentangling,menon2021longtail,provost2000machine,tang2020long,tian2020posterior,wu2021adversarial,zhang2021distribution} to adjust model outputs for the natural distribution. 

\textbf{Bias Correction.} We perform bias correction in two stages. After training has finished, we save a copy of all model parameter weights. We then fine-tune the model for a small fixed number of epochs on all data in the replay buffer using standard mini-batches (i.e., from the natural distribution). After fine-tuning, we perform class-specific bias correction on predicates for \gls{SPOP} questions and on attributes for \gls{SPAA} questions since they require a class as the answer. Correction is performed on these two question types independently.

The class-specific bias correction stage learns to adjust class-specific distances to the natural distribution. Our setup is reminiscent of Platt scaling~\cite{platt1999probabilistic}, which has been effective for model calibration~\cite{guo2017calibration}. Specifically, we compute network predictions for all \gls{SPOP} or \gls{SPAA} questions. We then compute target embeddings for all predicates or attributes in the class dictionary. We then compute the negative Euclidean distance of each predicted embedding to all target embeddings, which yields a score for each class. Finally, we train two parameters per class, $\alpha$ and $\beta$, to correct each class score: $s \leftarrow \alpha s + \beta$. These parameters are trained by minimizing a cross-entropy loss between the corrected score and the true label. After these stages, we evaluate the model and reset the model parameters back to their values from before bias correction. Resetting the parameters allows the model to perform better on balanced distributions and only use bias correction for evaluation in imbalanced settings.

\begin{figure}[t]
\begin{center}
  \includegraphics[width=0.6\linewidth]{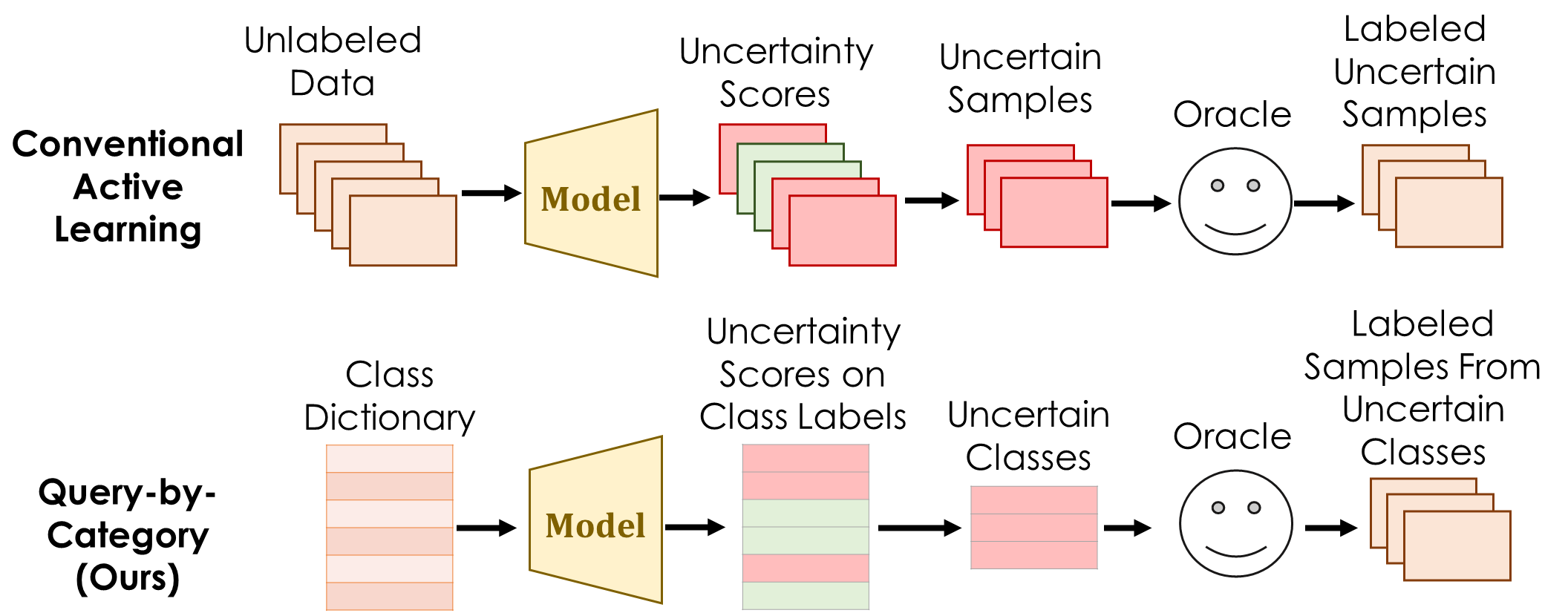}
\end{center}
\vspace{0.1in}
  \caption{
In classical active learning, a model computes uncertainty scores for unlabeled examples and an oracle labels uncertain examples. In our framework, a model computes uncertainty scores for \emph{classes} and an oracle provides examples from uncertain classes.\label{fig:framework}
}
\end{figure}

\subsection{Query-by-Category Framework}
\label{sec:tail-sampling}

A challenge in using conventional active sampling methods is that they rely on the following assumptions: they are being applied on a balanced set of classes, performance is measured using accuracy on the test data, and they ask for the label of a specific example. We argue that long-tailed distributions are more common in the real-world and agents should have the ability to ask questions beyond asking for just a label. To this end, we propose the Query-by-Category (QBCat) active learning framework that allows a learner to ask for examples from specific classes (Fig.~\ref{fig:framework}). To query an oracle for $P$ samples, the protocol is: 1) provide a dictionary of attribute and predicate classes to the learner; 2) the learner computes an uncertainty score for each class; 3) the learner uses weighted random sampling with class uncertainty scores as weights to select the class distribution for the $P$ samples; 4) the learner queries an oracle for $P$ samples using the class distribution from 3); and 5) the provided samples are combined with replay data and the model is updated. See Alg.~\ref{alg:active-overview} for an overview.

Using this framework, we propose a simple active sampling method called \textbf{QBCat-Tail} that prioritizes rare classes. This method assumes that a pre-training phase occurred on head classes of the dataset. During active learning, it assigns class uncertainty scores to tail classes \emph{uniformly at random}. An oracle then provides the learner with samples from the selected classes uniformly at random. While simple, classical active learning methods often do not consider the label distribution of the data to be learned, which can hinder their performance. 

\subsection{Model Architecture}
\label{sec:model}

\begin{figure}[t]
\begin{center}
  \includegraphics[width=0.85\linewidth]{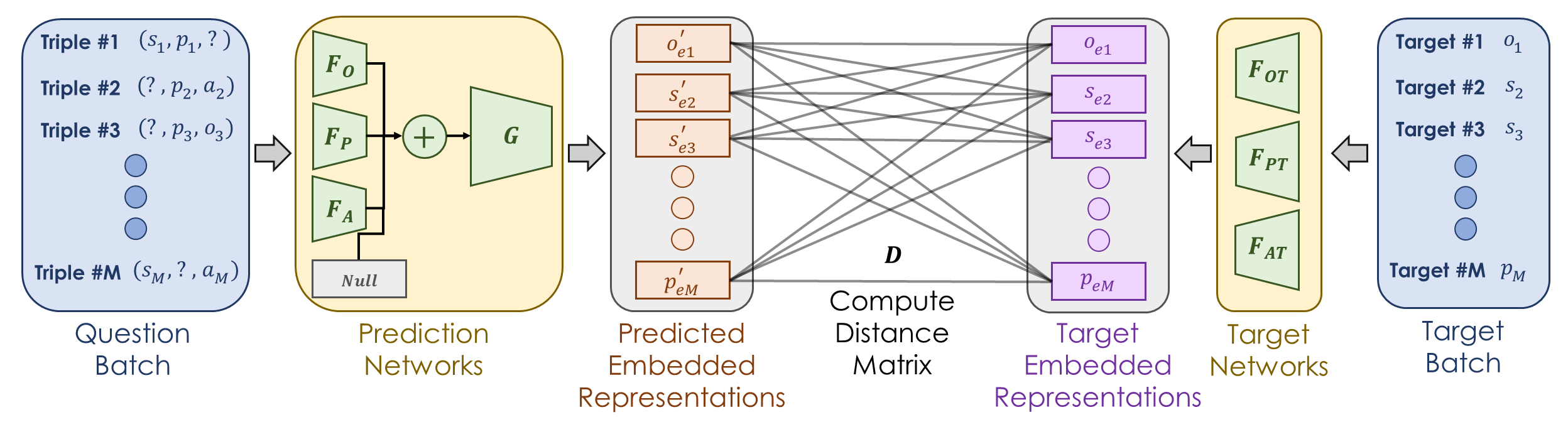}
\end{center}
\vspace{0.1in}
  \caption{
  Our architecture for a mini-batch with $M$ questions (triples) and targets. Question elements are processed by their respective embedding network/layer ($F_{O}$, $F_{P}$, $F_{A}$), concatenated with a trainable \texttt{Null}, and processed by $G$, which outputs \textit{predicted} embeddings. Targets are processed by their respective embedding network/layer ($F_{OT}$, $F_{PT}$, $F_{AT}$) to yield target embeddings. A cross-entropy loss is then computed between a pairwise distance matrix of all predicted and target embedding pairs ($D$) and a target identity matrix via Eq.~\ref{eq:loss}.
  }
\label{fig:model}
\end{figure}

Our model architecture takes an incomplete triple (question) as input and outputs a prediction for the missing element in the triple (see Fig.~\ref{fig:model}). Each subject, object, or predicate is represented as a vector. For subjects and objects that are bounding boxes, we ROI pool ground truth box features from a Faster R-CNN object detector~\cite{ren2015faster} pre-trained on MS-COCO~\cite{lin2014microsoft} (see Sec.~\ref{sec:model-details}). For predicates and objects that are strings, we use a lookup-table embedding.

Each triple contains two known elements and one missing element. The elements are either subject/object vectors $s,o \in \mathbb{R}^{d_{o}}$, a predicate label $p \in \mathbb{R}$, or an attribute label $a \in \mathbb{R}$. We define a feed-forward network $F_{O}: \mathbb{R}^{d_{o}} \rightarrow \mathbb{R}^{d}$ and two embedding layers, $F_{A}$ and $F_{P}$, to embed subjects/objects, attributes, or predicates in $d$-dimensional space respectively. We define one trainable $\texttt{N} \in \mathbb{R}^{d}$ (``Null'') per data type to represent the missing triple element, i.e., $\texttt{N}_{A}$, $\texttt{N}_{P}$, and $\texttt{N}_{O}$ represent a missing attribute, predicate, or subject/object respectively.

The two known elements in the triple are embedded in $d$-dimensional space using the corresponding embedding network/layers, i.e., $h_s=F_{O}\left(s\right)$, $h_o=F_{O}\left(o\right)$, $h_p=F_{P}\left(p\right)$, or $h_a=F_{A}\left(a\right)$. Then, the two embedded representations are concatenated with the appropriate trainable vector \texttt{N} to yield an embedded vector $h \in R^{3d}$, where \texttt{N} is concatenated in the exact location of the missing triple element, e.g., the question $q = \left(s,?,o\right)$ maps to $h=\left[h_s; \texttt{N}_{P}; h_o\right]$. This vector $h$ is processed by another feed-forward network $G: \mathbb{R}^{3d} \rightarrow \mathbb{R}^{d}$, which outputs a final predicted representation for the missing triple element. Similarly, we define a target embedding feed-forward network $F_{OT}$ and two target embedding layers $F_{AT}$ and $F_{PT}$ to embed target subject/objects, attributes, and predicates in $d$-dimensional space, respectively.

\textbf{Metric Learning Loss.} Given a mini-batch of $M$ questions, a predicted embedding $h_q$, its associated target embedding $t_q$, and all pairs between $h_u$ and $t_v$ in the mini-batch, we compute the categorical cross-entropy loss as:
\begin{equation}
\label{eq:loss}
    \mathcal{L} = -\log \left(\frac{\exp \left(-\|h_{q}-t_{q} \|^{2}\right)}{\sum_{u,v}\exp \left(-\|h_{u}-t_{v} \|^{2} \right)}\right) \enspace ,
\end{equation}
which encourages positive pairs to be embedded close to one another in feature space. This is equivalent to Neighborhood Component Analysis loss~\cite{goldberger2004neighbourhood}. Any distance function could be used, but Euclidean distance worked best in early experiments, so we use it here. 

\textbf{Model Inference.} We evaluate the model in three ways depending on if the answer is a subject/object, a predicate, or an attribute. First, we compute a predicted embedding for the question. When the target is a subject/object, we exploit the fact that the answer is limited to objects in the same image and compute target embeddings for each object in the image. For predicate/attribute targets, we compute target embeddings for all predicates/attributes. We then compute the negative Euclidean distance between the predicted embedding and all question-specific target embeddings to create a score vector for evaluation.

\section{Related Work}

\textbf{Link Prediction.} In link prediction, a neural network is provided with two objects and it predicts the relationship (``link'') between them. In our setting, a network performs link prediction between objects and node prediction of objects and attributes. Early link predictors were shallow networks that modeled relationships using simple algebraic operations~\cite{bordes2013translating,lin2015learning,nickel2011three,nickel2016holographic,socher2013reasoning,wang2014knowledge,yang2014embedding} (see \cite{nickel2015review} for a review). Recently, deep networks have demonstrated more success for modeling relationships due to their expressive power.
Knowledge Vault uses a multi-layer perceptron to learn from concatenated subject, relationship, and object embeddings~\cite{dong2014knowledge}. It has been applied to textual data. We extend Knowledge Vault to visual scenes due to its simplicity, which allows us to focus on active sampling.

\textbf{Long-Tailed Learning.} In long-tailed learning, methods tend to overfit to frequent classes and not generalize to rare classes. There are three main ways to train with imbalanced data (see \cite{zhang2021deep} for a survey): 1) re-balancing classes (e.g., via re-sampling~\cite{chawla2002smote,drumnond2003class,hu2020learning,mahajan2018exploring,shen2016relay}, loss adjustment~\cite{cao2019learning,cui2019class,khan2017cost,tan2020equalization}, or logit adjustment~\cite{hong2021disentangling,menon2021longtail,provost2000machine,tian2020posterior,tang2020long,wu2021adversarial,zhang2021distribution}), 2) transferring information from a pre-training stage or from more to less frequent classes~\cite{liu2019large,wang2019dynamic}, or 3) improving model performance via classifier design~\cite{kang2020decoupling,liu2020deep,wu2021adversarial} or ensembling~\cite{guo2021long,wang2020devil,zhou2020bbn}. Re-balancing strategies are simplest to implement and usually achieve comparable performance to more complex methods, so we use them here. Only a few long-tailed active learning approaches have been proposed~\cite{aggarwal2020active,bhattacharya2019generic,choi2021vab,gudovskiy2020deep}.
Similar to \cite{aggarwal2020active}, we modify the training setup to make methods more amenable to imbalanced data.

\textbf{Active Learning.} Active sampling methods attempt to select the fewest informative examples to be labeled by an oracle~\cite{cohn1994improving,lin2017active,settles2009active,wang2016cost,wei2015submodularity,yoo2019learning}.
Active learning strategies fall into three categories: model uncertainty~\cite{culotta2005reducing,dagan1995committee,dasgupta2008hierarchical,Lewis94asequential,scheffer2001active}, diversity-based sampling~\cite{gudovskiy2020deep,guo2010active,nguyen2004active,sener2018active}, and expected model change~\cite{freytag2014selecting,kading2016active,settles2007multiple,vezhnevets2012active}. Uncertainty can be quantified using entropy~\cite{holub2008entropy,luo2013latent,settles2008analysis}, discriminator scores~\cite{sinha2019variational}, margins between class probabilities~\cite{joshi2009multi,roth2006margin}, ensembling~\cite{beluch2018power,gal2017deep}, or model loss~\cite{yoo2019learning}. 
Uncertainty sampling is simple and easy to use with neural networks, so we use it here. Beyond this, there have been active sampling methods designed to discover rare classes and evaluated on standard classification tasks~\cite{he2008rare,hospedales2011finding,kading2015active}. While active learning has been widely explored for classification~\cite{beluch2018power,sener2018active,wang2016cost}, detection~\cite{feng2019deep,haussmann2020scalable,roy2018deep,sivaraman2014active}, and semantic segmentation~\cite{kasarla2019region,mackowiak2018cereals,siddiqui2020viewal,yang2017suggestive}, its exploration for node/link prediction has been limited~\cite{cai2017active,chen2014hallp,madhawa2020active,ostapuk2019activelink}. Most similar to our work are \cite{madhawa2020active,ostapuk2019activelink}, which meta-learn an uncertainty-based active sampler. Conversely, we are the first to perform active learning for node/link prediction on visual scenes.

\section{Experimental Setup}
\label{sec:experimental-protocol}

\subsection{Baselines}
\label{sec:active-sampling}

Active sampling is performed per question type, i.e., we select an equal number of samples from each question type from Sec.~\ref{sec:methods}. We compare four strategies that assign a weight of how likely each unlabeled sample is to be chosen: \textbf{Random} is simplest and uses uniform random weights. \textbf{Least Confident} computes the highest class score for an example, which is then inverted so examples with the smallest top scores are prioritized. \textbf{Minimum Margin} computes the margin between the highest and second highest score for an example~\cite{joshi2009multi,roth2006margin}. We invert this score such that samples with smaller margins receive larger weights. \textbf{Maximum Entropy} computes the entropy of softmaxed scores across classes~\cite{luo2013latent,settles2008analysis,yoo2019learning}.

We first assign a weight to each sample in the unlabeled dataset using one of these strategies. We then shift each weight, $w_{i}$, such that the smallest weight across samples is 1: $w_{i} \leftarrow w_{i} + (1-\min_{j}w_{j})$. Probabilities are then defined as $p_{i}=w_{i}/\sum_{j}w_{j}$ and weighted random sampling is used to select samples to be labeled by an oracle. We also compare two baselines. The pre-train baseline is trained on only pre-training data and evaluated immediately after. Pre-train is used to initialize the model for each active learner and is a lower bound. We also train an offline model on all training data, which is an upper bound. Both models are trained with standard mini-batches without bias correction.

\subsection{Visual Genome Dataset}

We conduct experiments on Visual Genome 1.4~\cite{krishnavisualgenome}. We pre-process the dataset following \cite{johnson2018image}, i.e., we partition the data into train (80\%), val (10\%), and test (10\%) sets, and filter objects based on size, number of occurrences, and number of relationships. This yields 62,565 train, 5,062 val, and 5,096 test images. We then form three questions from each relationship triple and each attribute triple in an image. This yields 3,474,969 train, 279,273 val, and 281,739 test questions (triples). The final dataset has 253 unique attribute classes (which includes object classes) and 46 unique predicate classes. All \gls{SPA} triples use the same predicate (``has attribute''). Histograms of attributes and predicates are in Fig.~\ref{fig:histograms}.

\subsection{Evaluation Protocol and Metrics}
\label{sec:evaluation-protocol}

We compute performance on the full test set, as well as a test set consisting of samples from the tail of the attribute and predicate distributions. We define head classes as those containing more samples than the mean across counts over all classes for predicates and attributes separately. This yields 66 head and 187 tail classes for attributes and 9 head and 37 tail classes for predicates. More details and class lists are in Sec.~\ref{sec:dataset-details}.

Given score vectors (from Sec.~\ref{sec:model}) and one-hot encodings of the answer, we compute two metrics for each question type: area under the receiver operating characteristic (AUROC) curve and mean average precision (mAP). When a subject/object is the target, we use sample-wise averaging across questions since images do not contain a uniform number of objects, i.e., score vectors are different lengths. When a predicate or attribute is the target, we use micro averaging. mAP emphasizes positive classes, making it more ideal for long-tailed datasets. 
It is also useful to summarize performance over increments. To do this, we define performance of an offline upper bound as $\gamma_{\text{offline}}$. Given the performance of a learner at increment $t$ as $\gamma_{t}$, overall performance is given by: $\Omega = \frac{1}{T}\sum_{t=1}^{T}{\left[1-\left(\gamma_{\text{offline}} - \gamma_{t}\right)\right]}$ over $T$ increments. If the learner performed as well as the offline method, then $\Omega=1$. Higher $\Omega$ values are better. $\Omega$ makes comparisons across question types, metrics, and test sets easier.

\section{Results}
\label{sec:main-results}

\begin{figure}[t]
\begin{center}
  \includegraphics[width=0.35\linewidth]{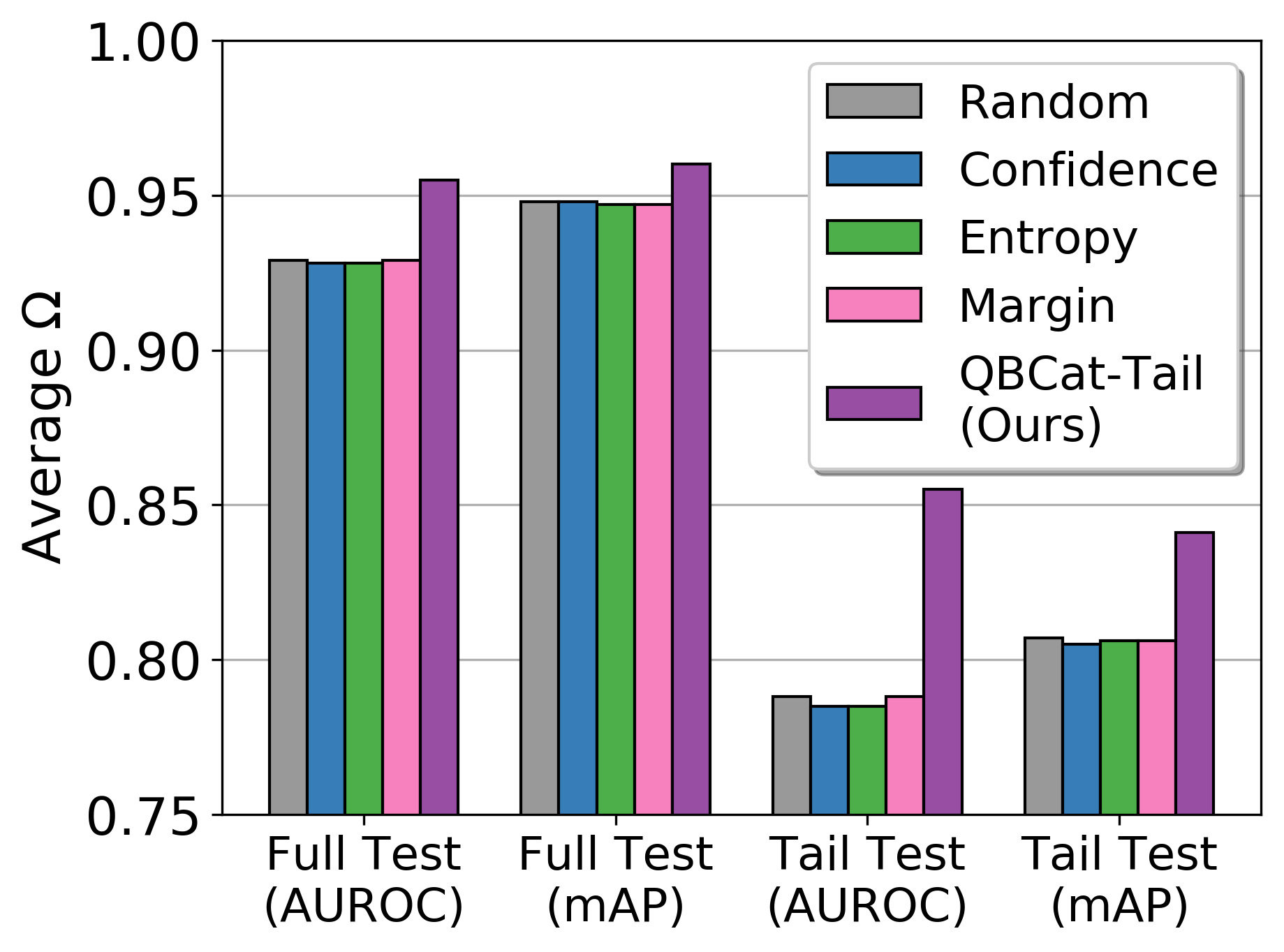}
\end{center}
\vspace{0.1in}
  \caption{
Average $\Omega$ performance of active learners over all 10 increments and five question types evaluated on the full and tail test sets. \label{fig:summary-chart}
}
\end{figure}

Each experiment was conducted with 10 random network initializations and we report the average over runs. After pre-training, all methods performed incremental active learning over 10 increments, where each active sampling method chose 100 samples from each of the six question types at each increment (i.e., 600 selected samples total per increment). Additional implementation details and hyper-parameters are in Sec.~\ref{sec:method-details}. While \gls{SPAP} questions are used during training, we do not report performance on them since it is uninteresting, i.e., it only requires the model to output the predicate ``has attribute.'' Our QBCat-Tail method uses the class breakdown from Fig.~\ref{fig:histograms} for determining which classes belong to the tail.

In Fig.~\ref{fig:summary-chart}, we plot the $\Omega$ scores of each method averaged over all five question types on each test set using AUROC and mAP. Raw $\Omega$ scores are in Table~\ref{tab:main-results}. When evaluated on the tail test set, our QBCat-Tail method outperforms baselines by a large margin. For example, it outperforms the closest baselines by 6.7\% in average $\Omega$ AUROC and 3.4\% in average $\Omega$ mAP. While performance differences are smaller on the full test set, our QBCat-Tail method outperforms the closest baselines by 2.6\% and 1.2\% in average $\Omega$ AUROC and mAP, respectively. The main advantage of our method is its ability to achieve strong performance on tail data without sacrificing performance on the natural data distribution. This strong performance could be a result of training on a balanced distribution of head and tail data, which could reduce over-fitting on head data. By reducing over-fitting, the model could be learning more generalized representations that improve overall performance.

\begin{figure*}[t]
    \centering
    \includegraphics[width=0.85\linewidth]{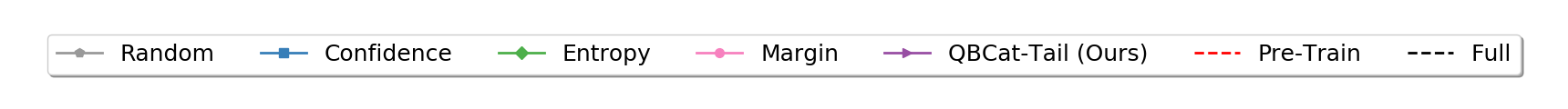}\\
    \centering
    \begin{subfigure}[t]{0.16\linewidth}
        \includegraphics[width=\linewidth]{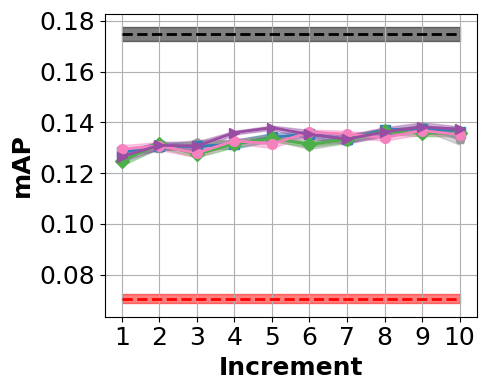}
        \caption{Full Test\\\quad\gls{SPAA}}
        \label{fig:main-results-spaa-subset}
    \end{subfigure} %
    \centering
    \begin{subfigure}[t]{0.16\linewidth}
        \includegraphics[width=\linewidth]{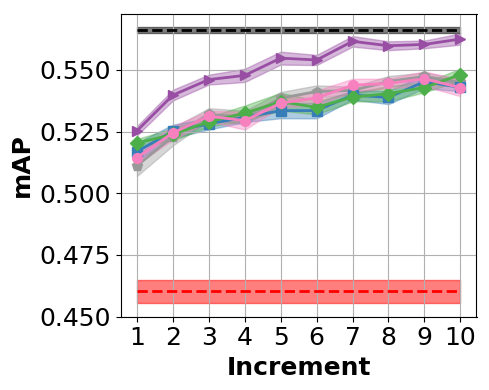}
        \caption{Full Test\\\quad\gls{SPOS}}
        \label{fig:main-results-spos-subset}
    \end{subfigure} %
    \centering
    \begin{subfigure}[t]{0.16\linewidth}
        \includegraphics[width=\linewidth]{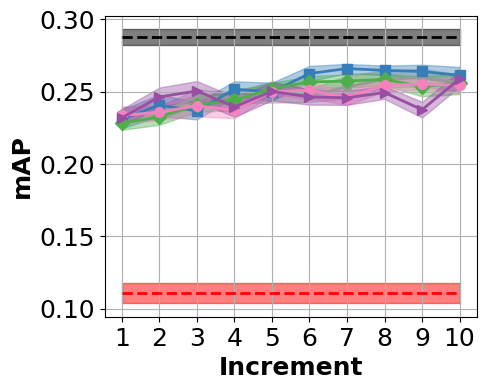}
        \caption{Full Test\\\quad\gls{SPOP}}
        \label{fig:main-results-spop-subset}
    \end{subfigure} %
    \centering
    \begin{subfigure}[t]{0.16\linewidth}
        \includegraphics[width=\linewidth]{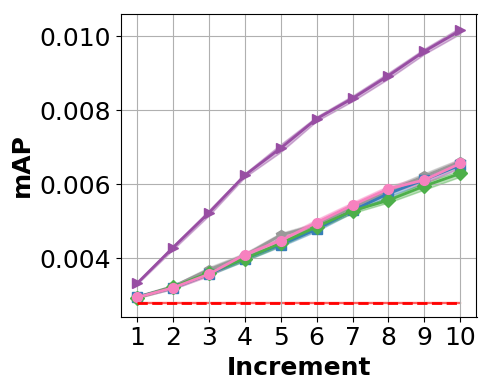}
        \caption{Tail Test\\\quad\gls{SPAA}}
        \label{fig:main-results-spaa-tail-subset}
    \end{subfigure} %
    \centering
    \begin{subfigure}[t]{0.16\linewidth}
        \includegraphics[width=\linewidth]{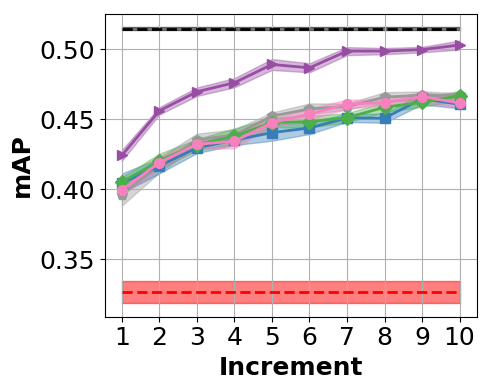}
        \caption{Tail Test\\\quad\gls{SPOS}}
        \label{fig:main-results-spos-tail-subset}
    \end{subfigure} %
    \centering
    \begin{subfigure}[t]{0.16\linewidth}
        \includegraphics[width=\linewidth]{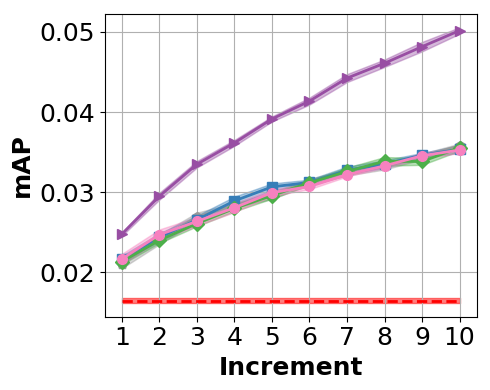}
        \caption{Tail Test\\\quad\gls{SPOP}}
        \label{fig:main-results-spop-tail-subset}
    \end{subfigure} %
    \vspace{0.1in}
    \caption{Incremental performance on the full and tail test sets for various question types. Each curve is the average over 10 runs with shaded standard error. For clarity, the offline upper bound has been removed from the tail plots for $\left(s, p, a?\right)$ and $\left(s, ?, o\right)$, where the offline baseline achieved an average mAP of 0.312 and 0.263 respectively. More plots are in Sec.~\ref{sec:additional-results}.
    }
    \label{fig:main-results-combined-subset}
\end{figure*}

To explore performance differences further, we show mAP learning curves for each active learning method in Fig.~\ref{fig:main-results-combined-subset}. Additional learning curves are in Sec.~\ref{sec:additional-results} and show similar trends. Across all increments, our QBCat-Tail method either rivals or outperforms alternative methods across question types and test sets. When evaluated on the full test set, our QBCat-Tail method outperforms all methods by a large margin on box-based questions (i.e., \gls{SPAS}, \gls{SPOS}, and \gls{SPOO}). For questions where the agent must predict a predicate \gls{SPOP} or an attribute \gls{SPAA}, performance of the QBCat-Tail method on the full dataset is similar to other active learning strategies. QBCat-Tail achieves significant performance improvements across question types on the tail test set. Overall, box-based questions appear to be easiest, yielding the highest performance values, while the \gls{SPOP} and \gls{SPAA} questions are harder, yielding the lowest performance values. This implies that it is more difficult for learners to predict specific attribute or predicate classes. Performance differences among baseline active learners across test sets are minimal. This is because baseline methods do not explore the tail of the distribution and oversample head data (see histograms in Sec.~\ref{sec:additional-experiments}). Additional studies of active learning baselines on only tail data are in Sec.~\ref{sec:additional-experiments}. 

\subsection{Additional Studies}
\label{sec:additional-experiments}

Next, we study several components of our training procedure to identify which yield the most improvement. In Table~\ref{tab:additional-studies}, we study the $\Omega$ performance of QBCat-Tail in four settings: \textbf{Standard Mini-Batches} uses standard mini-batch construction, i.e., batches are sampled uniformly at random without replacement. \textbf{Without Bias Correction} uses re-balanced mini-batches without bias correction. \textbf{Main Setup} uses re-balanced mini-batches, bias correction, and selects tail classes uniformly at random. \textbf{Frequency Probabilities} assigns each tail class a probability equal to $\frac{num\_samples\_in\_class}{num\_samples\_in\_dataset}$ with re-balanced mini-batches and bias correction.


\begin{table}[t]
\caption{$\Omega$ mAP performance for various versions of our QBCat-Tail method. Main Setup denotes the main model that uses re-balanced mini-batches, bias correction, and uniform class probabilities. Each result is the average over 10 runs.
    \vspace{0.1in}
\label{tab:additional-studies}}
\centering
\footnotesize
\begin{tabular}{lccccc}
\toprule
\textsc{Study} & \gls{SPAS} & \gls{SPAA} & \gls{SPOS} & \gls{SPOP} & \gls{SPOO} \\ 
\midrule
\multicolumn{2}{l}{\textit{Full Test Set}}\\
Without Bias Correction & 0.864 & 0.871 & 0.941 & 0.780 & 0.941 \\
Standard Mini-Batches & 0.870 & 0.952 & 0.968 & 0.919 & 0.955 \\
Main Setup & 0.919 & 0.959 & 0.985 & 0.958 & 0.979 \\
Frequency Probabilities & \textbf{0.921} & \textbf{0.962} & \textbf{0.990} & \textbf{0.973} & \textbf{0.981} \\
\midrule
\multicolumn{2}{l}{\textit{Tail Test Set}}\\
Without Bias Correction & 0.770 & 0.692 & 0.948 & \textbf{0.796} & 0.916 \\
Standard Mini-Batches & 0.669 & 0.691 & 0.921 & 0.756 & 0.894 \\
Main Setup & 0.819 & 0.695 & 0.965 & 0.776 & 0.949 \\
Frequency Probabilities & \textbf{0.828} & \textbf{0.696} & \textbf{0.971} & 0.783 & \textbf{0.954} \\
\bottomrule
\end{tabular}
\end{table}

On the full test set, performing bias correction is critical to model performance. Without bias correction, performance on \gls{SPOP} and \gls{SPAA} questions drops by 17.8\% and 8.8\% respectively. On both test sets, using standard mini-batches yields slightly worse performance than using re-balanced batches, with performance on \gls{SPAS} questions being the most negatively impacted. Surprisingly, performing bias correction improves performance on the tail classes for all question types except \gls{SPOP}. Frequency-based probabilities yield slightly better performance than uniform probabilities with a maximum performance difference of 1.5\% across questions and test sets. Future work could explore more methods for assigning tail probabilities (e.g., using class counters, pseudo-labeling, etc.).

\section{Discussion and Conclusion}

In this work, we introduced a new active learning framework and demonstrated its effectiveness on a new problem domain (active learning for visual triple completion). Specifically, we proposed new methods and training paradigms for incremental active learning of long-tailed attributes and relationships. We then introduced the Query-by-Category (QBCat) active learning setup which changes the framing of how agents ask oracles for more training data. It allows models to ask for an example of a particular class instead of asking for the label of an uncertain example. We then proposed the QBCat-Tail method and showed that when combined with suitable re-biasing, it performs comparably to existing active learning methods on the natural long-tailed distribution, and demonstrates significant performance improvements on tail classes. These contributions provide future researchers with a new active learning protocol, as well as a framework for studying how agents should ask for labels.

In the future, it would be interesting to extend to settings where more than one element in a triple (or a higher order relation) is missing, or to relations spanning multiple images. While we focused on uncertainty-based active learning methods due to their simplicity, it would be interesting to explore diversity-based methods that could exploit the metric space learned by our model. We demonstrated that asking for specific classes improved performance over standard active sampling methods. It would be interesting to develop methods that exploit this finding to improve tail performance further. Overall, our framing encourages future studies on how agents should pose questions to an oracle to best improve performance.

\ifthenelse{\boolean{ack}}{
\paragraph{Acknowledgements.} TH and CK were supported in part by the DARPA/SRI Lifelong Learning Machines program [HR0011-18-C-0051], NSF award \#1909696, and NSF award \#2047556. The views and conclusions contained herein are those of the authors and should not be interpreted as representing the official policies or endorsements of any sponsor. We thank Robik Shrestha, James Arnold, and Manoj Acharya for their comments and useful discussions.
}

\bibliography{egbib}

\ifthenelse{\boolean{combined}}{
\clearpage
\begin{center}
    {\Large Supplemental Material \normalsize}
\end{center}
\beginsupplement

\section{Algorithmic Overview of the Query-by-Category Active Learning Protocol}
\label{sec:algorithmic-overview}

As basic research, this paper is focused on the Query-by-Category framework and understanding its potential benefits over conventional active learning. Conventional active learning assumes an oracle that annotates selected examples between learning rounds. Here, we assume an oracle provides a list of classes to be identified. The algorithm then requests samples from specific classes in each round, and the oracle then provides inputs containing those classes.

A high-level overview of the Query-by-Category active learning protocol is in Alg.~\ref{alg:active-overview}. To query an oracle for $P$ samples, the protocol is as follows: 1) provide a dictionary of attribute and predicate classes to the learner; 2) the learner computes an uncertainty score for each class; 3) the learner uses weighted random sampling with class uncertainty scores as weights to select the class distribution for the $P$ samples; 4) the learner queries an oracle for $P$ samples using the class distribution from 3); and 5) the provided samples are combined with replay data and the model is updated. Note that this setup requires an initial class dictionary to be provided to the agent. This class dictionary could be initialized using an existing dataset and a human annotator could add more classes to the dictionary over time as scene conditions or objects change.

\RestyleAlgo{ruled}

\SetKwComment{Comment}{/* }{ */}

\setcounter{algocf}{0}
\renewcommand{\thealgocf}{S\arabic{algocf}}

\begin{algorithm}[h]
\caption{Query-by-Category Active Learning Protocol}\label{alg:active-overview}
\KwData{Dictionary of classes ($\mathcal{C}$); Dictionary of question types ($\mathcal{Q}$); Number of active samples to query per increment per question type ($P$), Replay buffer containing all pre-training data ($\mathcal{R}$)}
\KwResult{Updated model}
\While{increment}{
    Initialize empty dictionary $\mathcal{U}$ for class uncertainty scores\;
    \For{$c$ in $\mathcal{C}$}{
        Compute uncertainty score $s$ for class $c$\;
        Store uncertainty score: $\mathcal{U}[c] \gets s$\;
    }
    Initialize empty active learning data buffer $\mathcal{B}$\;
    \For{$q$ in $\mathcal{Q}$}{
        Sample $P$ classes randomly using class uncertainty scores $\mathcal{U}$ as weights\;
            \For{$c$ in $P$}{
                Query oracle for example from class $c$ of question type $q$\;
                Add new example to buffer $\mathcal{B}$\;
            }
    }
    Update model with data from $\mathcal{B}$ and $\mathcal{R}$\;
    Add data from $\mathcal{B}$ to replay buffer $\mathcal{R}$\;
}
\end{algorithm}

\subsection{QBCat-Tail Active Sampling}

Using the Query-by-Category active learning framework, we propose the QBCat-Tail active sampling method that assigns class uncertainty scores to tail classes uniformly at random. In a deployed setting, one way the QBCat-Tail method could identify samples as belonging to tail classes is by keeping a count of how many times each class in the class dictionary is visited. Those classes which have been rarely visited would be considered tail classes and assigned a uniform random probability of being selected. Those classes which have been frequently visited would be considered head classes and assigned a probability of zero. Note that QBCat-Tail is just one way of assigning class uncertainty scores and alternative ways of assigning scores using the QBCat framework is an area for future work.

\section{Method Details} 
\label{sec:method-details}

We include additional details related to our training paradigm, model architecture, and implementation details in the following subsections.

\subsection{Cross Validation}
\label{sec:cross-valid}
To perform cross-validation, we first use stratified random sampling to split the experience replay buffer and newly labeled samples into $k$ separate folds each, where $k$ is a hyper-parameter. Then, we combine $(k-1)$ folds from the experience replay buffer and new samples together to form a training set, and use the remaining held-out sets as a validation set. We train the model on the training set and compute validation loss each epoch. Once the validation loss has not improved for a pre-defined number of epochs (patience), we record the epoch where the model achieved the best validation loss and end cross-validation. While cross-validation is traditionally performed in $k$ separate rounds, we found that using a single round of training/validation was sufficient for determining the optimal number of epochs and reduces compute time, so we use this approach.

\subsection{Re-Balanced Mini-Batches} 
\label{sec:re-balance}

Since our active sampling methods select very few examples to be labeled in each increment, there is a large imbalance between old samples in the experience replay buffer and newly labeled examples. Further, there is a large imbalance between samples from more frequently and less frequently represented classes due to the long-tailed nature of the training data. To deal with these data imbalances, we perform an epoch in the following way. We iterate over newly labeled examples by selecting a fixed number $\sfrac{M}{2}$ at each iteration to be included in a mini-batch of size $M$. After we have iterated over all new data, we shuffle the new data and begin another epoch.

Simultaneously we iterate over all data in the experience replay buffer. For each batch selected from the replay buffer, we first perform hard negative mining to determine which pairs of samples are the most difficult in the batch.
Given a batch of replay samples, we perform hard negative mining in the following way. We first find all samples in the mini-batch where the correct answer is not in the top-$\ell$ predictions output by the network and pair the question example with its top predicted incorrect answer.
These hard negative pairs are then added to a buffer of negatives that is updated each iteration. We use a buffer of negatives to keep track of positive/negative pairs that the model struggles with throughout the increment to ensure that the model maintains performance on previous data. 

After hard negative mining, we randomly select $\sfrac{M}{4}$ pairs of negatives from the buffer to be included in the mini-batch with new examples (i.e., $\sfrac{M}{2}$ total old samples). After we combine the new samples with the negatives, we update the model for a single iteration. Once we have iterated over the entire experience replay buffer, we shuffle the replay buffer data and begin iterating again. We empty the hard negative buffer at the end of each active learning increment. Since our cross-validation procedure faces the same imbalance problem of many old samples to very few new samples, we use this same re-balanced mini-batch selection procedure to form balanced batches during both the training and validation stages of the cross-validation training stage.
We compare this re-balanced mini-batch selection process (which presents an equal number of new and old examples to the model in each batch) to standard mini-batch creation, i.e., uniform random sampling over a combination of new and old data, in our experiments (see Fig.~\ref{fig:additional-studies} and Sec.~\ref{sec:additional-study-standard-mb}).

\subsection{Model Training}

Here, we describe the model training procedure from Sec.~\ref{sec:model} in more detail. Given a mini-batch of $M$ questions and associated targets, we train the model using metric learning. More specifically, for the $M$ questions, we compute the $M$ predicted representations in embedding space and their associated $M$ target representations. We then compute pairwise distances between all combinations of question and target representations using the Euclidean distance metric. That is, we compute a distance matrix $D \in R^{M \times M}$ such that values on the diagonal of the matrix indicate the distance between a predicted embedding representation and the true associated target representation. Values on the off-diagonal indicate distances between unassociated question/target pairs. This formulation allows us to treat distances between true questions and targets as positives during training, while distances between unassociated questions and targets are treated as negatives. Formally, given a mini-batch of $M$ questions and targets, we compute the loss for a positive pair as follows. First, we compute a question embedding $h_{q}$ and its associated target embedding $t_{q}$. Then, given all batch pairs between question embeddings and target embeddings located at indices $u$ and $v$ respectively, we compute a categorical cross-entropy loss for the positive pair as:
\begin{equation}
\label{eq:loss-supp}
    \mathcal{L} = - \log \left(\frac{\exp \left(-\|h_{q}-t_{q} \|^{2}\right)}{\sum_{u, v}\exp \left(-\|h_{u}-t_{v} \|^{2} \right)}\right) \enspace ,
\end{equation}
which encourages positive pairs to be embedded closer to one another in feature space. This formulation is equivalent to Neighborhood Component Analysis loss~\cite{goldberger2004neighbourhood}. While any distance function could be used, we found Euclidean distance worked best in early experiments, so we use it here.

\subsection{Implementation Details}
\label{sec:model-details}

All feed-forward neural networks in the model architecture use the same network consisting of two layers with 256 units in the first layer and 128 units in the second layer. The first layer is a fully-connected layer with batch norm and a Mish activation function~\cite{misra2019mish}, which helps prevent gradient vanishing. The second layer is the same as the first, but replaces the Mish activation with a sigmoid activation such that all output vector entries are between zero and one, which is useful when computing Euclidean distances between vectors.

For cross-validation and full model training, models are trained using stochastic gradient descent with a learning rate of 0.01, a weight decay factor of $10^{-5}$, a momentum value of 0.9, and a re-balanced mini-batch size of 512 (i.e., 256 new samples and 256 old samples). Before selecting 256 old samples to be included in the mini-batch, we use a batch size of 800 samples from the replay buffer to perform hard negative mining. For choosing the number of training epochs, we use a cross-validation $k$ value of 5, a cross-validation patience of 10, a validation batch size of 512, and set a maximum limit of 100 epochs. For hard negative mining, we use a top-$\ell$ value of 3. For the first stage of bias correction, we use the Adam optimizer and train for 10 epochs with a learning rate of 0.01 and cosine annealing. For the second stage of bias correction, we use the LBFGS optimizer and train for 500 iterations with a learning rate of 0.01.

For pre-training models, offline upper bound models, and experiments using standard mini-batches, we use a batch size of 256. We pre-train on 2,500 samples randomly selected from each head predicate class (minus ``has attribute'') and head attribute class, which results in 185,000 pre-training samples. We found pre-training for 1 epoch was sufficient for model convergence and using more epochs for pre-training caused overfitting to the head classes. We train the offline upper bounds for 25 epochs.

To encode subjects and objects as vectors, we first pre-train a Faster R-CNN object detection model~\cite{ren2015faster} with a ResNet-50 backbone on the MS COCO dataset~\cite{lin2014microsoft}. We then use the Faster R-CNN model to extract 1024-dimensional feature vectors after the ROI pooling layer for all ground truth object boxes in all images. Since we are focused on the node/link prediction task rather than object detection, we assume access to ground truth boxes; however, future work could explore node/link prediction performance when using boxes generated from the region proposal network of the object detection network. We train the Faster R-CNN model from torchvision using the following hyper-parameters:
backbone=ResNet-50, optimizer=stochastic gradient descent, learning rate=0.02, learning rate decays by a factor of 10 at epochs 16 and 22, momentum=0.9, aspect ratio group factor=3, batch size=2, data augmentation=horizontal flips, epochs=26, weight decay=0.0001, across 8 GPUs. After training, this model achieves an average precision value (at Intersection over Union (IoU) of 0.5) of 50.7\% when evaluated on the COCO mini-val set. In all experiments, we use pre-annotated scene graphs from the Visual Genome dataset to form our dataset triples.

\section{Dataset Details}
\label{sec:dataset-details}

To partition attribute and predicate categories into head and tail classes, we do the following. For attributes, we first compute the number of samples for each class represented in the training dataset. We then compute the mean across counts of all classes, which is equal to 10,973 samples. We then define head attribute classes as those classes containing more than 10,000 samples in the training set and tail attribute classes as those containing fewer than 10,000 samples. Similarly, for predicates, we first compute the number of samples for each class in the training dataset, with the exception of the ``has attribute'' predicate. The mean across predicate counts is equal to 15,525 samples. Head predicate classes are then defined as those classes containing more than 15,000 samples and tail predicate classes are defined as those classes containing fewer than 15,000 samples. This yields 66 head attribute classes, 187 tail attribute classes, 9 head predicate classes (which includes ``has attribute''), and 37 tail predicate classes. Histograms of the counts of all training samples for attribute and predicate classes sorted from smallest to largest are in Fig.~\ref{fig:histograms}, along with the associated test histograms. Note that both the train and test distributions are long-tailed. We provide lists of the exact head and tail classes for attributes and predicates below, which are sorted from most to least frequently represented in the training dataset.

\begin{figure*}[t]
    \begin{subfigure}[t]{0.5\linewidth}
        \centering
        \includegraphics[width=\linewidth]{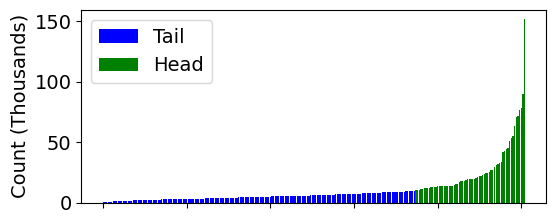}
        \caption{Train Attributes}
        \label{fig:attribute-distribution}
    \end{subfigure} %
    \begin{subfigure}[t]{0.5\linewidth}
        \centering
        \includegraphics[width=\linewidth]{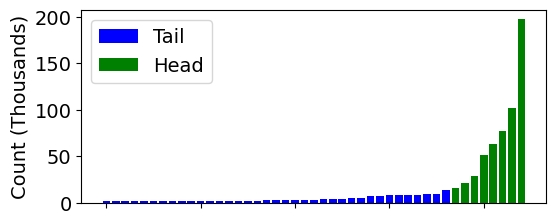}
        \caption{Train Predicates}
        \label{fig:predicate-distribution}
    \end{subfigure}
    \\
    \begin{subfigure}[t]{0.5\linewidth}
        \centering
        \includegraphics[width=\linewidth]{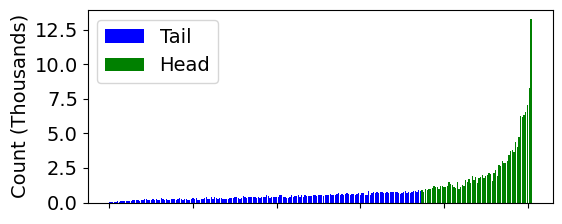}
        \caption{Test Attributes}
        \label{fig:attribute-distribution-test}
    \end{subfigure} %
    \begin{subfigure}[t]{0.5\linewidth}
        \centering
        \includegraphics[width=\linewidth]{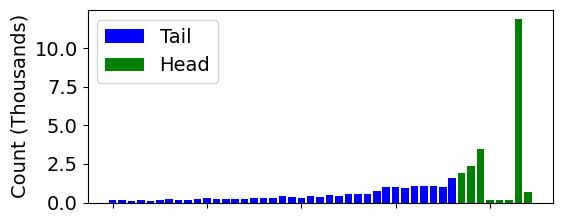}
        \caption{Test Predicates}
        \label{fig:predicate-distribution-test}
    \end{subfigure}
    \vspace{0.1in}
    \caption{Histograms of attribute and predicate train and test distributions.
    }
    \label{fig:histograms}
\end{figure*}

\paragraph{Attribute Head Classes:}
\begin{verbatim}
    [white, black, tree, man, green, blue, brown, shirt,
      wall, building, window, sky, red, ground, head,
      grass, person, large, woman, hair, table, leg,
      yellow, cloud, sign, gray, car, wooden, pant, grey,
      fence, hand, water, chair, shadow, small, floor,
      tall, door, jacket, leaf, road, line, plate, long,
      field, sidewalk, arm, dark, standing, background,
      people, boy, clear, face, street, snow, metal, ear,
      bush, short, girl, pole, orange, light, bag]
\end{verbatim}

\paragraph{Attribute Tail Classes:} 
\begin{verbatim}
    [here, tan, track, shoe, jean, glass, bus, picture,
       tile, sitting, plant, train, wheel, pillow, branch,
       bench, giraffe, rock, silver, tire, umbrella, roof,
       tail, pink, wood, dirt, stripe, horse, elephant,
       short, flower, big, food, boat, dog, parked, zebra,
       coat, hat, bowl, box, hill, mountain, reflection,
       neck, brick, wave, cloudy, cabinet, walking, young,
       round, striped, bike, house, trunk, open, counter,
       helmet, top, cat, handle, mirror, foot, glass,
       board, bed, motorcycle, back, clock, ceiling, cow,
       bottle, animal, truck, curtain, beach, frame, sand,
       banana, shelf, paper, seat, bear, bird, cup, photo,
       lady, purple, pizza, ocean, smiling, bare, sheep,
       lamp, plastic, windshield, blonde, part, empty, wire,
       skateboard, nose, old, child, wing, letter, book,
       player, container, looking, wet, railing, kite,
       design, plane, stand, basket, sink, edge, wood,
       ski, surfboard, bright, towel, brick, cap, logo,
       beige, post, writing, finger, vehicle, playing,
       concrete, stone, hanging, glove, orange, dirty, calm,
       boot, engine, tie, pot, spot, apple, light, little,
       colorful, flag, glasses, mouth, grassy, square, dry,
       thick, painted, paw, gold, closed, shiny, word,
       sock, thin, stone, one, light brown, part, leather,
       distant, flying, eye, on, ball, headlight,
       rectangular, sticker, number, horn, hole, sunglass,
       sliced, button, knob, key, tennis]
\end{verbatim}

\paragraph{Predicate Head Classes:}
\begin{verbatim}
    [has attibute, on, has, in, wearing, of, behind, with,
       near]
\end{verbatim}

\paragraph{Predicate Tail Classes:}
\begin{verbatim}
    [next to, on top of, holding, by, under,
       in front of, wears, above, sitting on, standing on,
       beside, riding, on side of, standing in, over, at,
       walking on, attached to, around, hanging on, covering,
       below, sitting in, eating, carrying, laying on,
       against, have, parked on, for, along, looking at,
       belonging to, inside, and, made of, covered in]
\end{verbatim}

\section{Additional Results}
\label{sec:additional-results}

Here, we include a table with baseline model performances as well as raw $\Omega$ scores for our main experiments from Sec.~\ref{sec:main-results}. We then include learning curves of our main results using the mAP and AUROC metrics. We then provide tables containing the raw $\Omega$ scores for several additional studies involving standard mini-batches, the performance of active learning methods using re-balanced mini-batches without bias correction, and the performance of active learning methods when selecting samples from only tail attribute classes and tail predicate classes. Additionally, we include histogram distributions of the number of samples selected from each attribute class and predicate class by each active learning method.

\subsection{Main Results Tables}

\begin{table*}[t]
\caption{Accuracy of the Pre-Train and Offline baselines on each question type evaluated on the full and tail test sets. Each result is the average over 10 runs and used to compute the $\Omega$ metric.
    \vspace{0.1in}
\label{tab:baseline-results}}
\centering
\resizebox{\linewidth}{!}{
\begin{tabular}{lcccccccccc}
\toprule
& \multicolumn{5}{c}{\textsc{\textbf{AUROC}}} & \multicolumn{5}{c}{\textsc{\textbf{mAP}}} \\
\cmidrule(r){2-6} \cmidrule(r){7-11}
\textsc{Baseline} & \gls{SPAS} & \gls{SPAA} & \gls{SPOS} & \gls{SPOP} & \gls{SPOO} & \gls{SPAS} & \gls{SPAA} & \gls{SPOS} & \gls{SPOP} & \gls{SPOO}  \\ 
\midrule
\multicolumn{2}{l}{\textit{Full Test Set}}\\
Pre-Train & 0.716 & 0.769 & 0.691 & 0.710 & 0.663 & 0.483 & 0.071 & 0.460 & 0.111 & 0.400 \\
Offline & 0.878 & 0.951 & 0.794 & 0.913 & 0.776 & 0.675 & 0.175 & 0.566 & 0.288 & 0.510 \\
\midrule
\multicolumn{2}{l}{\textit{Tail Test Set}}\\
Pre-Train & 0.515 & 0.368 & 0.565 & 0.402 & 0.558 & 0.279 & 0.003 & 0.327 & 0.016 & 0.311 \\
Offline & 0.895 & 0.960 & 0.759 & 0.906 & 0.783 & 0.708 & 0.312 & 0.515 & 0.263 & 0.532 \\
\bottomrule
\end{tabular}
}
\end{table*}

Table~\ref{tab:baseline-results} contains raw accuracies for the Pre-Train and Offline baseline models across question types and test sets. The Offline accuracy values are used for normalization with $\Omega$.

Table~\ref{tab:main-results} contains the raw $\Omega$ scores for each active learning method across question types and test sets for our main experiments from Sec.~\ref{sec:main-results}. Note that these are the raw $\Omega$ scores that are averaged across question types to generate Fig.~\ref{fig:summary-chart}. QBCat-Tail represents the main version of our Tail method that selects classes uniformly at random, while QBCat-Tail (Freq.) represents the additional study of the QBCat-Tail method from Sec.~\ref{sec:additional-experiments} that selects classes with probabilities defined by the class frequencies. While the QBCat-Tail (Freq.) method performs the best consistently across metrics, question types, and test sets, QBCat-Tail performs comparably to QBCat-Tail (Freq.) without needing access to class frequencies. QBCat-Tail outperforms or performs comparably to baseline active learning methods when evaluated on the full test set and outperforms baseline methods by a significant margin when evaluated on the tail test set. However, on average, QBCat-Tail outperforms all baseline methods, as shown in Fig.~\ref{fig:summary-chart}.

\begin{table*}[t]
\caption{$\Omega$ performance of each active learning method over all 10 increments on each question type evaluated on the full and tail test sets. We report performance using both the AUROC and mAP metrics to compute $\Omega$. Each result is the average over 10 runs. QBCat-Tail is our main method that selects classes with uniform random probabilities, while QBCat-Tail (Freq.) selects classes with probabilities defined by the class frequencies. Each method was run using \textbf{re-balanced mini-batches and bias correction}.
    \vspace{0.1in}
\label{tab:main-results}}
\centering
\resizebox{\linewidth}{!}{
\begin{tabular}{lcccccccccc}
\toprule
& \multicolumn{5}{c}{\textsc{\textbf{AUROC}}} & \multicolumn{5}{c}{\textsc{\textbf{mAP}}} \\
\cmidrule(r){2-6} \cmidrule(r){7-11}
\textsc{Model} & \gls{SPAS} & \gls{SPAA} & \gls{SPOS} & \gls{SPOP} & \gls{SPOO} & \gls{SPAS} & \gls{SPAA} & \gls{SPOS} & \gls{SPOP} & \gls{SPOO}  \\ 
\midrule
\multicolumn{2}{l}{\textit{Full Test Set}}\\
Random & 0.917 & 0.890 & 0.965 & 0.915 & 0.958 & 0.894 & 0.958 & 0.969 & 0.961 & 0.959 \\
Confidence & 0.918 & 0.889 & 0.961 & 0.917 & 0.955 & 0.894 & 0.959 & 0.967 & 0.965 & 0.957 \\
Entropy & 0.917 & 0.888 & 0.964 & 0.915 & 0.956 & 0.893 & 0.957 & 0.969 & 0.960 & 0.957 \\
Margin & 0.918 & 0.890 & 0.964 & 0.917 & 0.956 & 0.894 & 0.958 & 0.969 & 0.958 & 0.957 \\
QBCat-Tail & 0.941 & 0.927 & 0.978 & 0.951 & 0.974 & 0.919 & 0.959 & 0.985 & 0.958 & 0.979 \\
QBCat-Tail (Freq.) & \textbf{0.943} & \textbf{0.934} & \textbf{0.983} & \textbf{0.964} & \textbf{0.976} & \textbf{0.921} & \textbf{0.962} & \textbf{0.990} & \textbf{0.973} & \textbf{0.981} \\
\midrule
\multicolumn{2}{l}{\textit{Tail Test Set}}\\
Random & 0.792 & 0.581 & 0.929 & 0.723 & 0.915 & 0.740 & 0.693 & 0.931 & 0.766 & 0.904 \\
Confidence & 0.794 & 0.576 & 0.921 & 0.725 & 0.910 & 0.742 & 0.693 & 0.925 & 0.767 & 0.902 \\
Entropy & 0.791 & 0.574 & 0.927 & 0.724 & 0.911 & 0.740 & 0.693 & 0.928 & 0.766 & 0.902 \\
Margin & 0.795 & 0.582 & 0.927 & 0.724 & 0.911 & 0.743 & 0.693 & 0.929 & 0.766 & 0.901 \\
QBCat-Tail & 0.866 & 0.695 & 0.959 & 0.804 & 0.951 & 0.819 & 0.695 & 0.965 & 0.776 & 0.949 \\
QBCat-Tail (Freq.) & \textbf{0.872} & \textbf{0.716} & \textbf{0.966} & \textbf{0.836} & \textbf{0.955} & \textbf{0.828} & \textbf{0.696} & \textbf{0.971} & \textbf{0.783} & \textbf{0.954} \\
\bottomrule
\end{tabular}
}
\end{table*}

\subsection{Additional Plots for Main Results}

\begin{figure*}[t]
    \centering
    \includegraphics[width=0.85\linewidth]{images/final_tail_results_rebalanced_mb_legend.png}\\
    \centering
    \begin{subfigure}[t]{0.19\linewidth}
        \includegraphics[width=\linewidth]{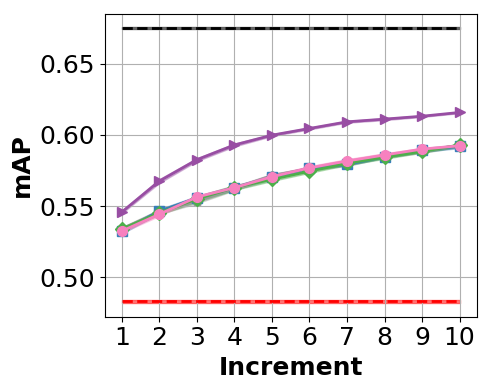}
        \caption{\gls{SPAS}}
        \label{fig:main-results-spas}
    \end{subfigure} %
    \centering
    \begin{subfigure}[t]{0.19\linewidth}
        \includegraphics[width=\linewidth]{images/final_full_results_rebalanced_mb_mAP_SPAA.png}
        \caption{\gls{SPAA}}
        \label{fig:main-results-spaa}
    \end{subfigure} %
    \centering
    \begin{subfigure}[t]{0.19\linewidth}
        \includegraphics[width=\linewidth]{images/final_full_results_rebalanced_mb_mAP_SPOS.png}
        \caption{\gls{SPOS}}
        \label{fig:main-results-spos}
    \end{subfigure} %
    \centering
    \begin{subfigure}[t]{0.19\linewidth}
        \includegraphics[width=\linewidth]{images/final_full_results_rebalanced_mb_mAP_SPOP.png}
        \caption{\gls{SPOP}}
        \label{fig:main-results-spop}
    \end{subfigure} %
    \centering
    \begin{subfigure}[t]{0.19\linewidth}
        \includegraphics[width=\linewidth]{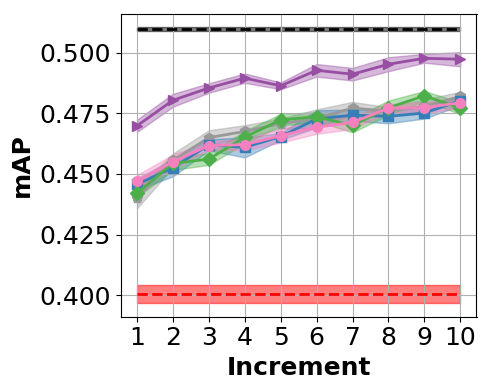}
        \caption{\gls{SPOO}}
        \label{fig:main-results-spoo}
    \end{subfigure} %
    \\
    \centering
    \begin{subfigure}[t]{0.19\linewidth}
        \includegraphics[width=\linewidth]{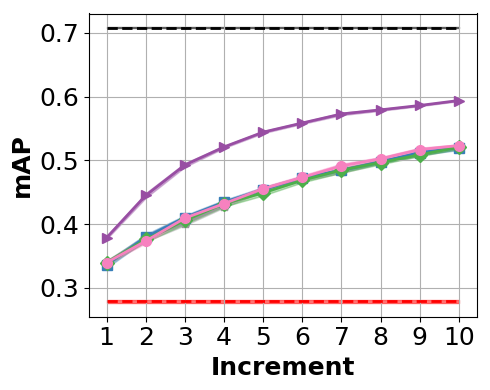}
        \caption{\gls{SPAS}}
        \label{fig:main-results-spas-tail}
    \end{subfigure} %
    \centering
    \begin{subfigure}[t]{0.19\linewidth}
        \includegraphics[width=\linewidth]{images/final_tail_results_rebalanced_mb_mAP_SPAA.png}
        \caption{\gls{SPAA}}
        \label{fig:main-results-spaa-tail}
    \end{subfigure} %
    \centering
    \begin{subfigure}[t]{0.19\linewidth}
        \includegraphics[width=\linewidth]{images/final_tail_results_rebalanced_mb_mAP_SPOS.png}
        \caption{\gls{SPOS}}
        \label{fig:main-results-spos-tail}
    \end{subfigure} %
    \centering
    \begin{subfigure}[t]{0.19\linewidth}
        \includegraphics[width=\linewidth]{images/final_tail_results_rebalanced_mb_mAP_SPOP.png}
        \caption{\gls{SPOP}}
        \label{fig:main-results-spop-tail}
    \end{subfigure} %
    \centering
    \begin{subfigure}[t]{0.19\linewidth}
        \includegraphics[width=\linewidth]{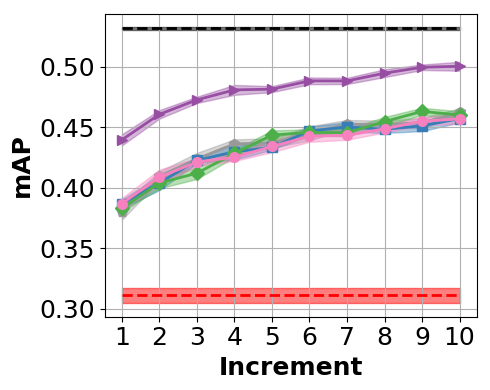}
        \caption{\gls{SPOO}}
        \label{fig:main-results-spoo-tail}
    \end{subfigure} %
    \vspace{0.1in}
    \caption{mAP learning curves showing incremental learning performance on the \textbf{full test set} (top) and \textbf{tail test set} (bottom) over 10 increments for each question type. We also include the performance of pre-train (lower bound) and full offline (upper bound) models. Each curve is the average over 10 runs and the standard error over runs is denoted by the shaded region. For plot clarity, the offline upper bound has been removed from the tail plots for $\left(s, p, a?\right)$ and $\left(s, ?, o\right)$, where the offline baseline achieved an average mAP of 0.312 and 0.263, respectively.
    }
    \label{fig:main-results-combined}
\end{figure*}
\begin{figure*}[t]
    \centering
    \includegraphics[width=0.85\linewidth]{images/final_tail_results_rebalanced_mb_legend.png}\\
    \centering
    \begin{subfigure}[t]{0.19\linewidth}
        \includegraphics[width=\linewidth]{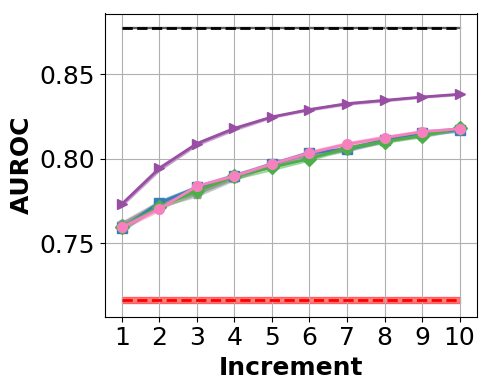}
        \caption{\gls{SPAS}}
        \label{fig:main-results-spas-auroc}
    \end{subfigure} %
    \centering
    \begin{subfigure}[t]{0.19\linewidth}
        \includegraphics[width=\linewidth]{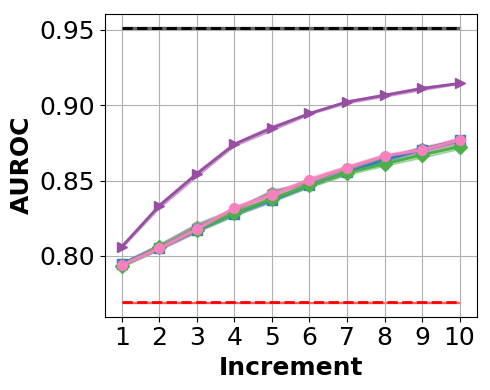}
        \caption{\gls{SPAA}}
        \label{fig:main-results-spaa-auroc}
    \end{subfigure} %
    \centering
    \begin{subfigure}[t]{0.19\linewidth}
        \includegraphics[width=\linewidth]{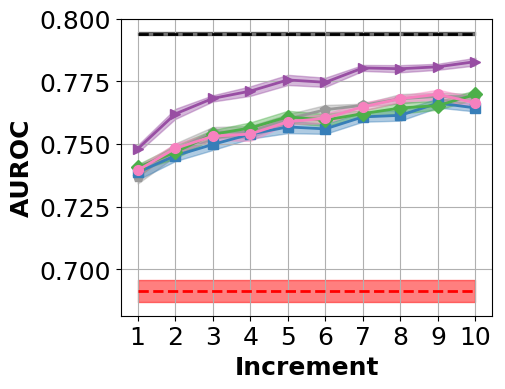}
        \caption{\gls{SPOS}}
        \label{fig:main-results-spos-auroc}
    \end{subfigure} %
    \centering
    \begin{subfigure}[t]{0.19\linewidth}
        \includegraphics[width=\linewidth]{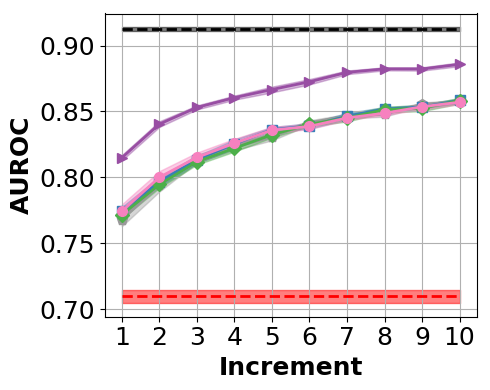}
        \caption{\gls{SPOP}}
        \label{fig:main-results-spop-auroc}
    \end{subfigure} %
    \centering
    \begin{subfigure}[t]{0.19\linewidth}
        \includegraphics[width=\linewidth]{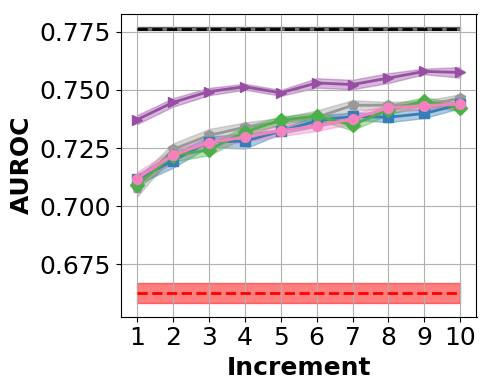}
        \caption{\gls{SPOO}}
        \label{fig:main-results-spoo-auroc}
    \end{subfigure} %
    \\
    \centering
    \begin{subfigure}[t]{0.19\linewidth}
        \includegraphics[width=\linewidth]{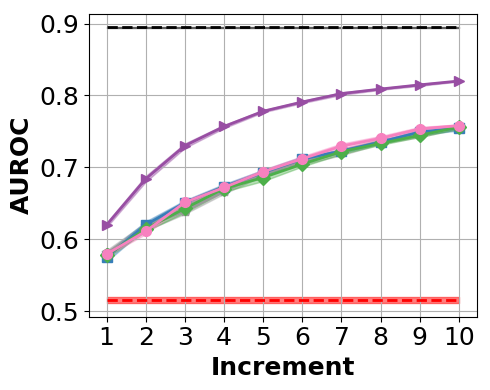}
        \caption{\gls{SPAS}}
        \label{fig:main-results-spas-tail-auroc}
    \end{subfigure} %
    \centering
    \begin{subfigure}[t]{0.19\linewidth}
        \includegraphics[width=\linewidth]{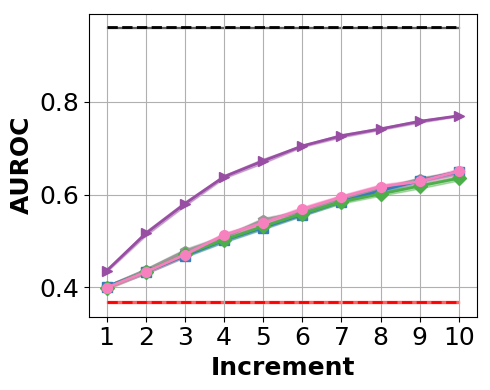}
        \caption{\gls{SPAA}}
        \label{fig:main-results-spaa-tail-auroc}
    \end{subfigure} %
    \centering
    \begin{subfigure}[t]{0.19\linewidth}
        \includegraphics[width=\linewidth]{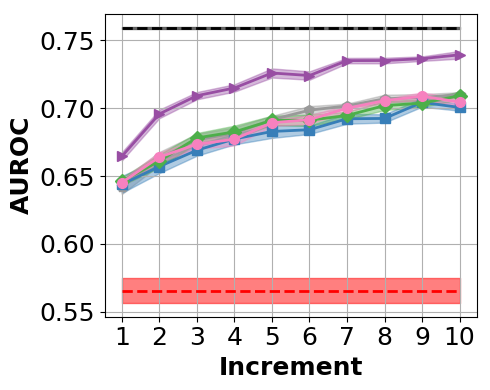}
        \caption{\gls{SPOS}}
        \label{fig:main-results-spos-tail-auroc}
    \end{subfigure} %
    \centering
    \begin{subfigure}[t]{0.19\linewidth}
        \includegraphics[width=\linewidth]{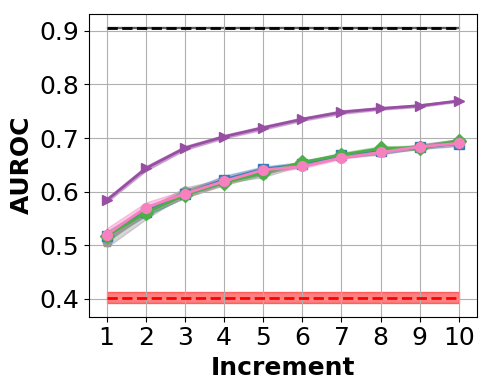}
        \caption{\gls{SPOP}}
        \label{fig:main-results-spop-tail-auroc}
    \end{subfigure} %
    \centering
    \begin{subfigure}[t]{0.19\linewidth}
        \includegraphics[width=\linewidth]{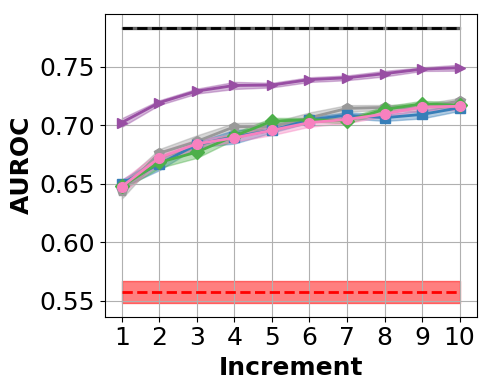}
        \caption{\gls{SPOO}}
        \label{fig:main-results-spoo-tail-auroc}
    \end{subfigure} %
    \vspace{0.1in}
    \caption{AUROC learning curves showing incremental learning performance on the \textbf{full test set} (top) and \textbf{tail test set} (bottom) over 10 increments for each question type. We also include the performance of pre-train (lower bound) and full offline (upper bound) models. Each curve is the average over 10 runs and the standard error over runs is denoted by the shaded region.
    }
    \label{fig:main-results-combined-auroc}
\end{figure*}

In Fig.~\ref{fig:main-results-combined-subset}, we showed a subset of learning curves using the mAP metric. In Fig.~\ref{fig:main-results-combined}, we show all learning curves using the mAP metric. In  Fig.~\ref{fig:main-results-combined-auroc}, we show the same learning curves using the AUROC metric. In mAP, our tail method outperforms or performs comparably to baselines on the full test set and outperforms all baselines by a large margin on the tail test set. In AUROC, our tail method outperforms all baselines by a large margin on all question types when evaluated on both the full test set and tail test set. Overall, baseline active learning methods perform similarly to the random sampling baseline on all question types on both test sets using both metrics.

\begin{figure}[t]
\begin{center}
  \includegraphics[width=0.5\linewidth]{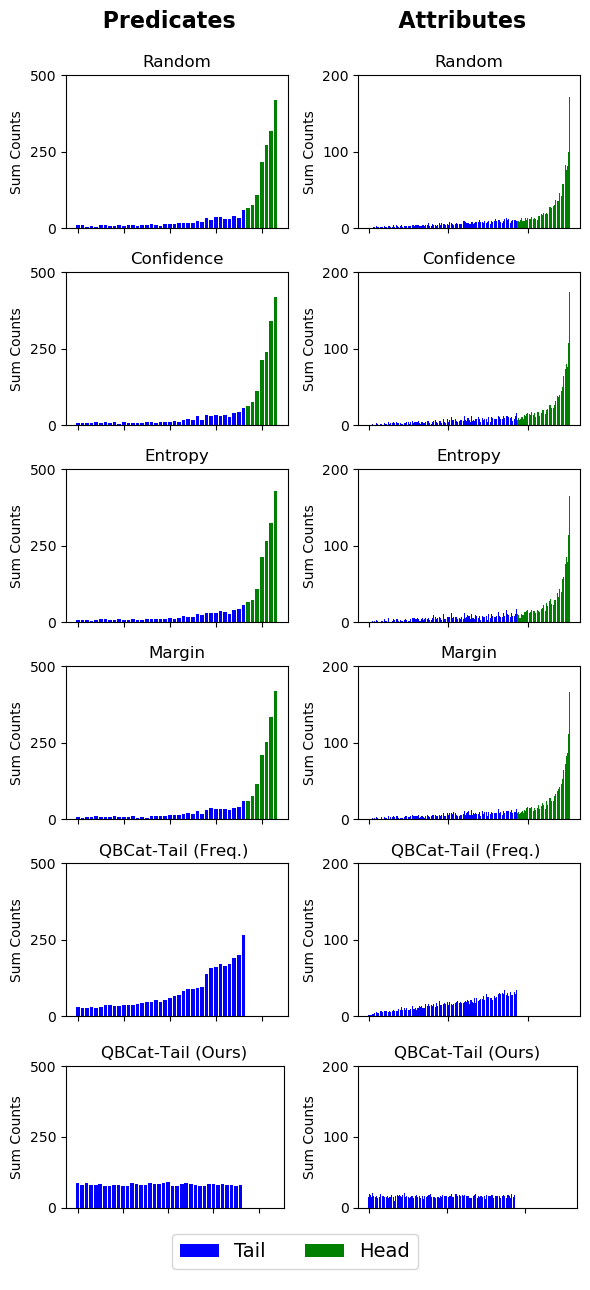}
\end{center}
\vspace{0.1in}
  \caption{Sum of counts of each predicate class and attribute class selected during active learning by each method after 10 increments. Each plot is averaged over 10 runs. QBCat-Tail (Ours) is our main method that selects classes with uniform random probabilities, while QBCat-Tail (Freq.) selects classes with probabilities defined by the class frequencies.
  }
\label{fig:sample-distributions}
\end{figure}

Since we perform active sampling at each increment, we also show histograms of the sum of samples selected from different predicate classes and attribute classes by each active learning method after all 10 increments in Fig.~\ref{fig:sample-distributions}. Note that we include the main variant of our Tail method that selects classes uniformly (QBCat-Tail), as well as the variant that selects classes based on their frequency (QBCat-Tail (Freq.)). Unsurprisingly, the baseline active sampling methods choose many samples from the head classes, which is likely the reason for their poor generalization to tail classes. Conversely, our tail-based sampling approach only selects samples from the tail of the distribution, allowing it to perform well on a wider variety of classes.

\subsection{Additional Studies Plots}

\begin{figure}[t]
    \centering
    \includegraphics[width=0.6\linewidth]{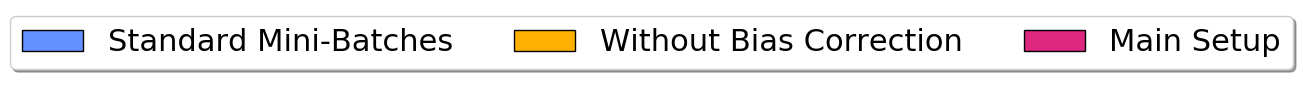}\\
    \centering
    \begin{subfigure}[t]{0.24\linewidth}
        \centering
        \includegraphics[width=\linewidth]{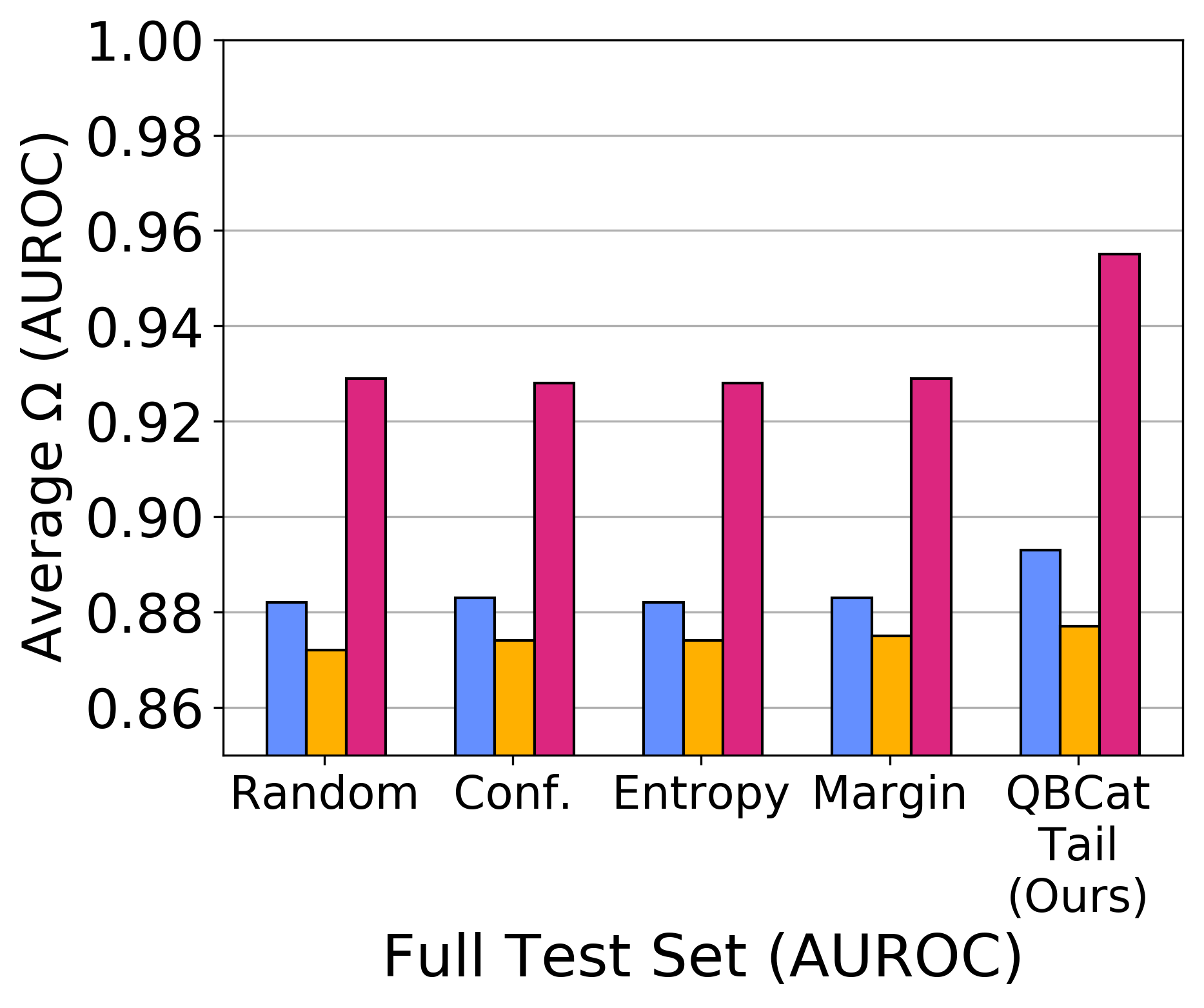}
        \label{fig:additional-study-auroc-full}
    \end{subfigure}
    \centering
    \begin{subfigure}[t]{0.24\linewidth}
        \centering
        \includegraphics[width=\linewidth]{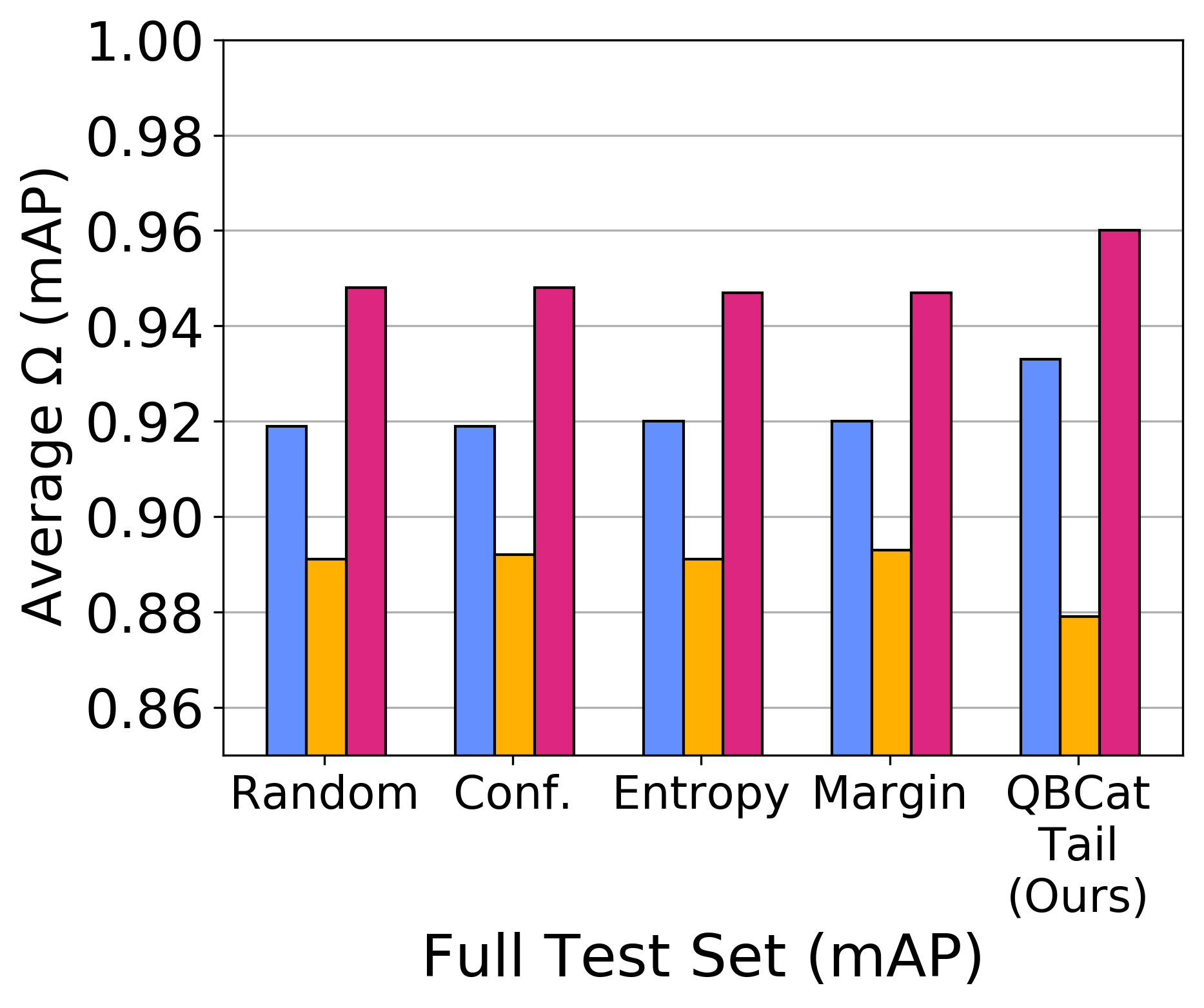}
        \label{fig:additional-study-map-full}
    \end{subfigure}
    \centering
    \begin{subfigure}[t]{0.24\linewidth}
        \centering
        \includegraphics[width=\linewidth]{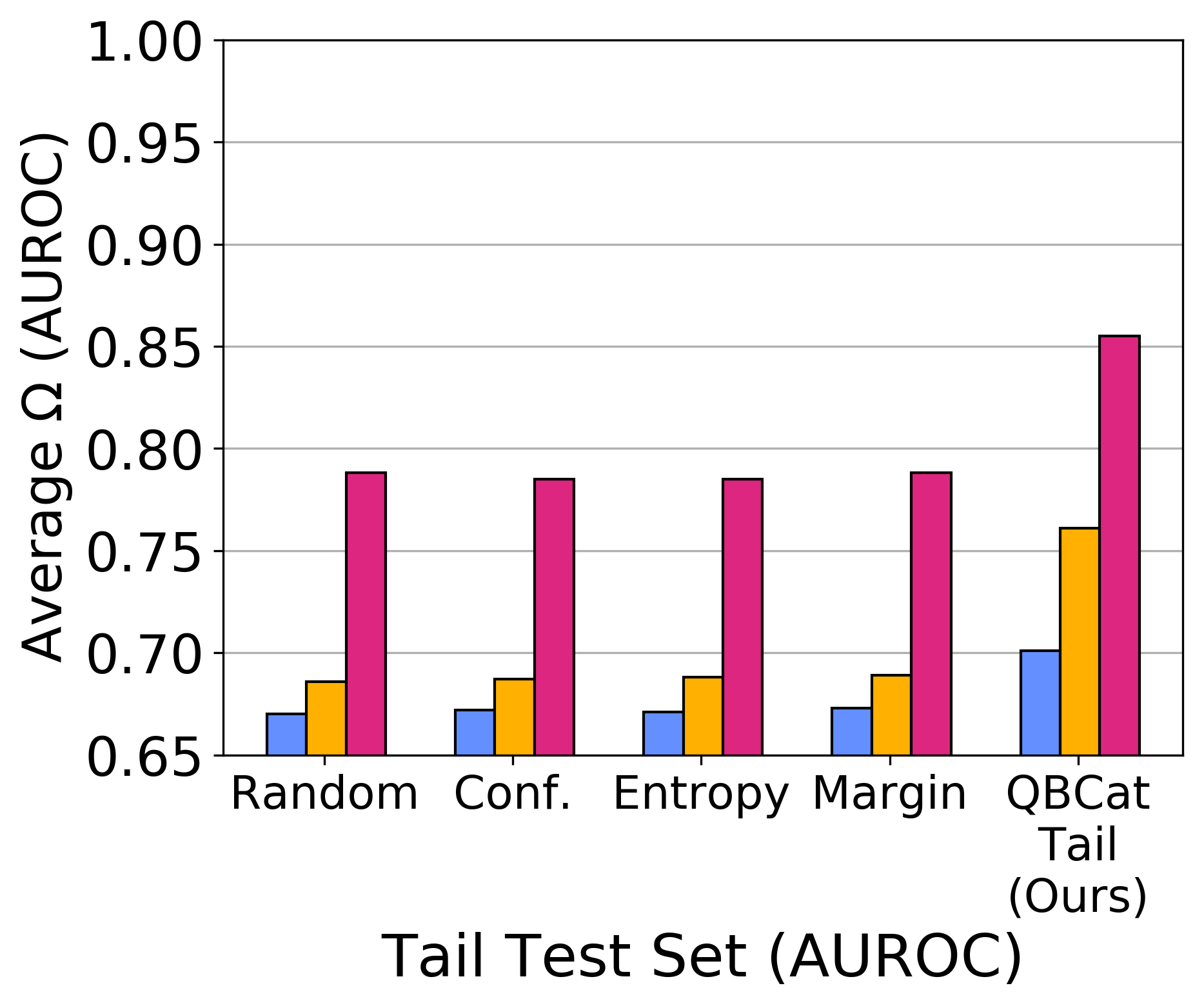}
        \label{fig:additional-study-auroc-tail}
    \end{subfigure}
    \centering
    \begin{subfigure}[t]{0.24\linewidth}
        \centering
        \includegraphics[width=\linewidth]{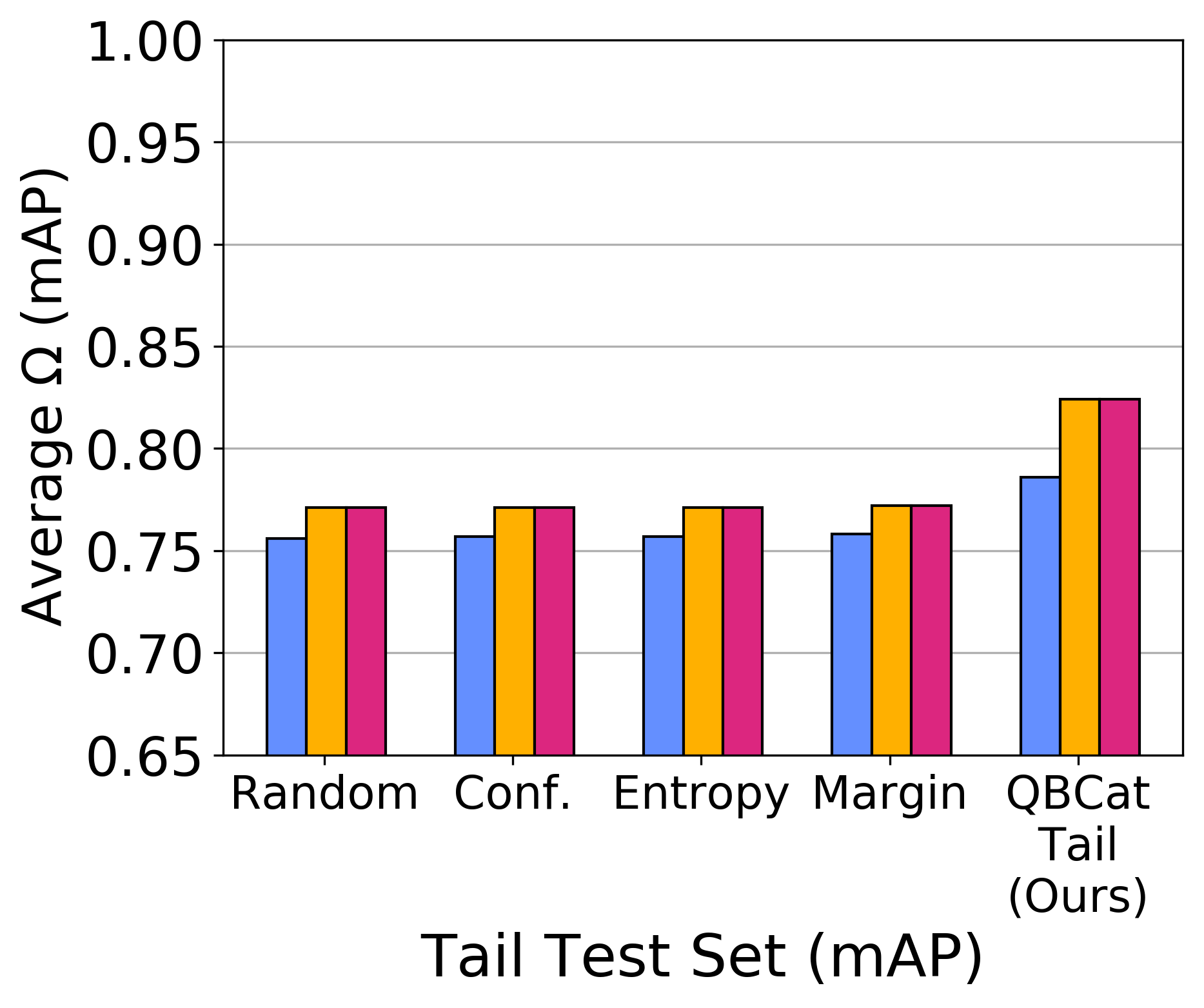}
        \label{fig:additional-study-map-tail}
    \end{subfigure}
    \vspace{0.1in}
    \caption{Average $\Omega$ performance of active learners under various setups on the full and tail test sets. Main Setup uses re-balanced mini-batches and bias correction. Each result is averaged over five question types.
    }
    \label{fig:additional-studies}
\end{figure}

In Fig.~\ref{fig:additional-studies}, we provide overview plots of our additional studies. In these plots, we compare performance of our main setup to using standard mini-batches and using re-balanced mini-batches without bias correction. Specifically, we plot the average $\Omega$ AUROC and average $\Omega$ mAP scores of each active learning method over all five question types. In the following subsections, we provide the raw $\Omega$ scores for each experiment.

On the full test set, we find that performing bias correction is critical to performance across models; however, our QBCat-Tail method is most affected by the absence of bias correction. This is expected as QBCat-Tail prioritizes data from tail classes and its weights must be readjusted for the natural long-tailed distribution via bias correction. On the tail test set, bias correction does not yield any benefit in terms of average $\Omega$ mAP, but yields benefit in average $\Omega$ AUROC. When evaluating on both the full and tail test sets, using standard mini-batches yields slightly worse performance than using re-balanced batches across methods. It is interesting to note that our QBCat-Tail method outperforms all other methods on both test sets when using standard mini-batches. However, all methods exhibit improved performance when trained using re-balanced mini-batches, motivating their need in long-tailed settings.

\subsection{Additional Studies with Standard Mini-Batches}
\label{sec:additional-study-standard-mb}

\begin{figure*}[th!]
    \centering
    \includegraphics[width=0.85\linewidth]{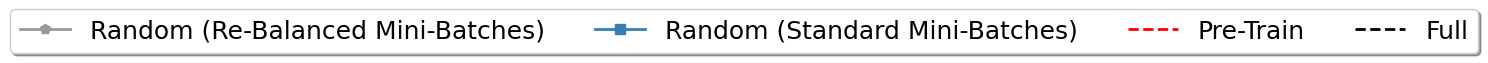}\\
    \centering
    \begin{subfigure}[t]{0.19\linewidth}
        \includegraphics[width=\linewidth]{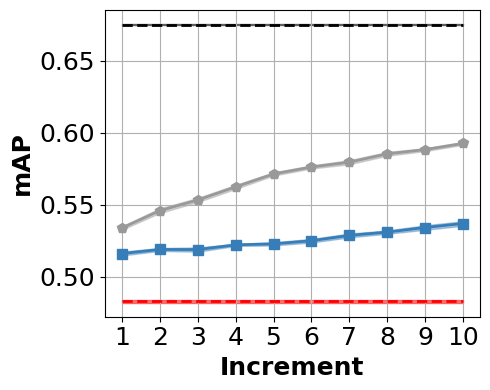}
        \caption{\gls{SPAS}}
        \label{fig:main-results-spas-standard-mb}
    \end{subfigure} %
    \centering
    \begin{subfigure}[t]{0.19\linewidth}
        \includegraphics[width=\linewidth]{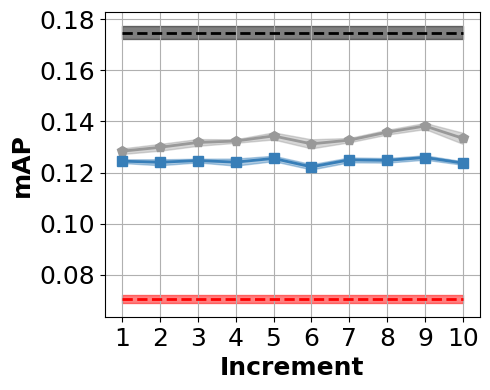}
        \caption{\gls{SPAA}}
        \label{fig:main-results-spaa-standard-mb}
    \end{subfigure} %
    \centering
    \begin{subfigure}[t]{0.19\linewidth}
        \includegraphics[width=\linewidth]{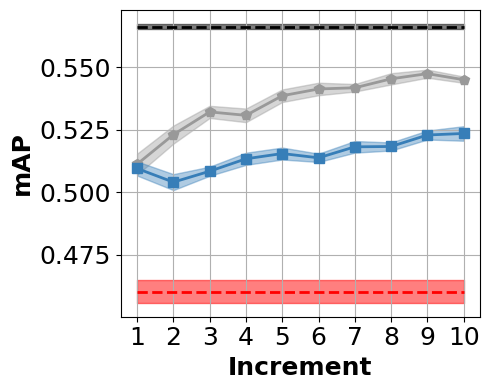}
        \caption{\gls{SPOS}}
        \label{fig:main-results-spos-standard-mb}
    \end{subfigure} %
    \centering
    \begin{subfigure}[t]{0.19\linewidth}
        \includegraphics[width=\linewidth]{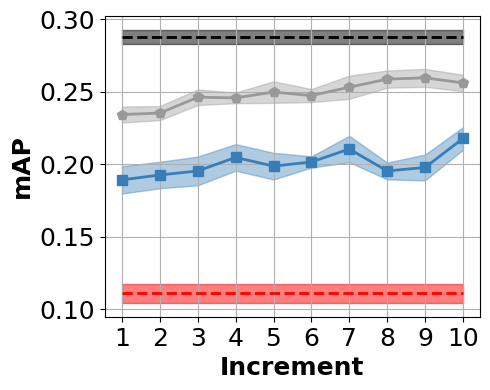}
        \caption{\gls{SPOP}}
        \label{fig:main-results-spop-standard-mb}
    \end{subfigure} %
    \centering
    \begin{subfigure}[t]{0.19\linewidth}
        \includegraphics[width=\linewidth]{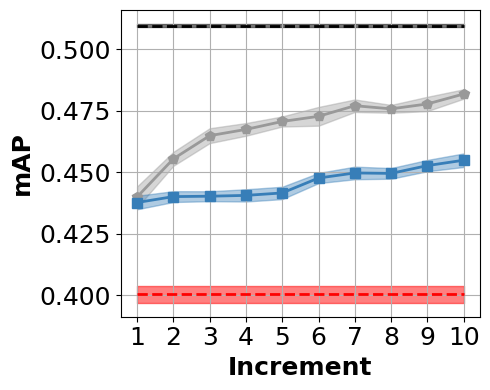}
        \caption{\gls{SPOO}}
        \label{fig:main-results-spoo-standard-mb}
    \end{subfigure} %
    \\
    \centering
    \begin{subfigure}[t]{0.19\linewidth}
        \includegraphics[width=\linewidth]{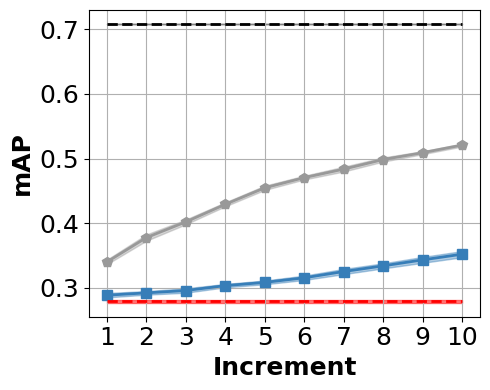}
        \caption{\gls{SPAS}}
        \label{fig:main-results-spas-tail-standard-mb}
    \end{subfigure} %
    \begin{subfigure}[t]{0.19\linewidth}
        \includegraphics[width=\linewidth]{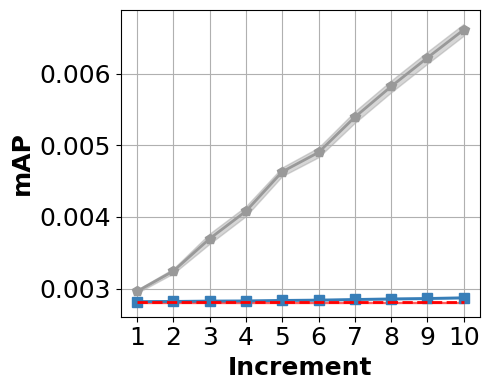}
        \caption{\gls{SPAA}}
        \label{fig:main-results-spaa-tail-standard-mb}
    \end{subfigure} %
    \centering
    \begin{subfigure}[t]{0.19\linewidth}
        \includegraphics[width=\linewidth]{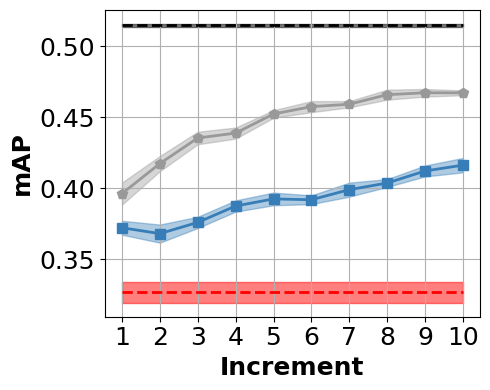}
        \caption{\gls{SPOS}}
        \label{fig:main-results-spos-tail-standard-mb}
    \end{subfigure} %
    \centering
    \begin{subfigure}[t]{0.19\linewidth}
        \includegraphics[width=\linewidth]{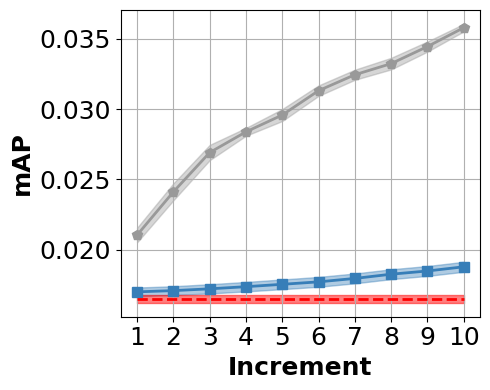}
        \caption{\gls{SPOP}}
        \label{fig:main-results-spop-tail-standard-mb}
    \end{subfigure} %
    \centering
    \begin{subfigure}[t]{0.19\linewidth}
        \includegraphics[width=\linewidth]{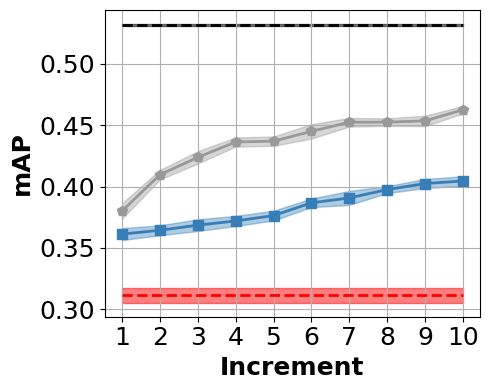}
        \caption{\gls{SPOO}}
        \label{fig:main-results-spoo-tail-standard-mb}
    \end{subfigure} %
    \vspace{0.1in}
    \caption{Learning curves comparing incremental learning performance of the random active sampling method using re-balanced mini-batches and standard mini-batches on the \textbf{full test set} (top) and \textbf{tail test set} (bottom) over 10 increments for each question type. We also include the performance of pre-train (lower bound) and full offline (upper bound) models. Each curve is the average over 10 runs and the standard error over runs is denoted by the shaded region. For plot clarity, the offline upper bound has been removed from the tail plots for $\left(s, p, a?\right)$ and $\left(s, ?, o\right)$, where the offline baseline achieved an average mAP of 0.312 and 0.263, respectively.
    }
    \label{fig:main-results-combined-standard-mb}
\end{figure*}

\begin{table*}[t]
\caption{$\Omega$ performance of each active learning method using \textbf{standard mini-batches} over all increments when evaluated on the full and tail test sets. We report performance on each question type using both the AUROC and mAP metrics to compute $\Omega$. Each result is the average over 10 runs and computed based on 10 increments.
    \vspace{0.1in}
\label{tab:main-results-standard-mb}}
\centering
\resizebox{\linewidth}{!}{
\begin{tabular}{lcccccccccc}
\toprule
& \multicolumn{5}{c}{\textsc{\textbf{AUROC}}} & \multicolumn{5}{c}{\textsc{\textbf{mAP}}} \\
\cmidrule(r){2-6} \cmidrule(r){7-11}
\textsc{Model} & \gls{SPAS} & \gls{SPAA} & \gls{SPOS} & \gls{SPOP} & \gls{SPOO} & \gls{SPAS} & \gls{SPAA} & \gls{SPOS} & \gls{SPOP} & \gls{SPOO}  \\
\midrule
\multicolumn{2}{l}{\textit{Full Test Set}}\\
Random & 0.875 & 0.834 & 0.944 & 0.822 & 0.936 & 0.850 & 0.950 & 0.949 & 0.913 & 0.936 \\
Confidence & 0.876 & \textbf{0.835} & 0.945 & \textbf{0.823} & 0.936 & 0.851 & 0.950 & 0.948 & 0.911 & 0.937 \\
Entropy & 0.876 & 0.834 & 0.946 & 0.821 & 0.934 & 0.851 & 0.950 & 0.950 & 0.913 & 0.935 \\
Margin & 0.876 & \textbf{0.835} & 0.948 & 0.822 & 0.936 & 0.851 & 0.950 & 0.952 & 0.911 & 0.936 \\
QBCat-Tail & \textbf{0.897} & 0.834 & \textbf{0.961} & 0.819 & \textbf{0.955} & \textbf{0.870} & \textbf{0.952} & \textbf{0.968} & \textbf{0.919} & \textbf{0.955} \\
\midrule
\multicolumn{2}{l}{\textit{Tail Test Set}}\\
Random & 0.663 & \textbf{0.415} & 0.878 & 0.534 & 0.863 & 0.607 & \textbf{0.691} & 0.877 & 0.754 & 0.851 \\
Confidence & 0.668 & \textbf{0.415} & 0.881 & 0.535 & 0.863 & 0.611 & \textbf{0.691} & 0.878 & 0.754 & 0.853 \\
Entropy & 0.666 & \textbf{0.415} & 0.882 & 0.531 & 0.860 & 0.610 & \textbf{0.691} & 0.879 & 0.754 & 0.850 \\
Margin & 0.665 & \textbf{0.415} & 0.887 & 0.533 & 0.863 & 0.610 & \textbf{0.691} & 0.886 & 0.754 & 0.851 \\
QBCat-Tail & \textbf{0.730} & \textbf{0.415} & \textbf{0.917} & \textbf{0.537} & \textbf{0.906} & \textbf{0.669} & \textbf{0.691} & \textbf{0.921} & \textbf{0.756} & \textbf{0.894} \\
\bottomrule
\end{tabular}
}
\end{table*}

In Sec.~\ref{sec:methods-imbalance}, we claimed that naive training using standard mini-batch construction (i.e., uniformly random sampled batches) caused models to overfit to head classes and impaired learning in later training increments. To support this claim, we compare the mAP performance of the random active sampling baseline using re-balanced mini-batches to using standard mini-batches in Fig.~\ref{fig:main-results-combined-standard-mb}. Overall, we see that performance using standard mini-batches is consistently lower than using re-balanced batches, especially when evaluated on the tail test set for \gls{SPAA} and \gls{SPOP} questions. This motivates the need for re-balanced mini-batches when training on long-tailed data distributions.

Further, we show the $\Omega$ performance in AUROC and mAP of each active learning method when using standard mini-batches in Fig.~\ref{fig:additional-studies} and Table~\ref{tab:main-results-standard-mb}. When using standard mini-batches, our QBCat-Tail method still performs comparably to or outperforms baseline active learning methods across question types and test sets. However, performance of all methods is improved by using re-balanced mini-batches, which further motivates their use when operating on imbalanced datasets.

\subsection{Additional Studies Without Bias Correction}

\begin{table*}[t]
\caption{$\Omega$ performance of each active learning method with re-balanced mini-batches \textbf{without bias correction} over all increments when evaluated on the full and tail test sets. We report performance on each question type using both the AUROC and mAP metrics to compute $\Omega$. Each result is the average over 10 runs and computed based on 10 increments.
\vspace{0.1in}
\label{tab:main-results-no-bias-correction}}
\centering
\resizebox{\linewidth}{!}{
\begin{tabular}{lcccccccccc}
\toprule
& \multicolumn{5}{c}{\textsc{\textbf{AUROC}}} & \multicolumn{5}{c}{\textsc{\textbf{mAP}}} \\
\cmidrule(r){2-6} \cmidrule(r){7-11}
\textsc{Model} & \gls{SPAS} & \gls{SPAA} & \gls{SPOS} & \gls{SPOP} & \gls{SPOO} & \gls{SPAS} & \gls{SPAA} & \gls{SPOS} & \gls{SPOP} & \gls{SPOO}  \\
\midrule
\multicolumn{2}{l}{\textit{Full Test Set}}\\
Random & 0.866 & \textbf{0.819} & 0.937 & 0.819 & 0.925 & 0.830 & 0.897 & 0.931 & 0.872 & 0.926 \\
Confidence & 0.867 & \textbf{0.819} & 0.936 & \textbf{0.822} & 0.926 & 0.831 & \textbf{0.898} & 0.929 & 0.877 & 0.926 \\
Entropy & 0.867 & \textbf{0.819} & 0.938 & 0.821 & 0.923 & 0.831 & \textbf{0.898} & 0.932 & 0.869 & 0.923 \\
Margin & 0.866 & \textbf{0.819} & 0.939 & \textbf{0.822} & 0.927 & 0.830 & \textbf{0.898} & 0.933 & \textbf{0.878} & 0.927 \\
QBCat-Tail & \textbf{0.898} & 0.807 & \textbf{0.944} & 0.804 & \textbf{0.934} & \textbf{0.864} & 0.871 & \textbf{0.941} & 0.780 & \textbf{0.941} \\
\midrule
\multicolumn{2}{l}{\textit{Tail Test Set}}\\
Random & 0.714 & 0.419 & 0.896 & 0.542 & 0.860 & 0.652 & 0.691 & 0.898 & 0.755 & 0.858 \\
Confidence & 0.717 & 0.419 & 0.892 & 0.547 & 0.862 & 0.655 & 0.691 & 0.894 & 0.755 & 0.860 \\
Entropy & 0.717 & 0.419 & 0.897 & 0.546 & 0.860 & 0.657 & 0.691 & 0.897 & 0.755 & 0.857 \\
Margin & 0.718 & 0.418 & 0.899 & 0.545 & 0.864 & 0.656 & 0.691 & 0.898 & 0.755 & 0.861 \\
QBCat-Tail & \textbf{0.824} & \textbf{0.443} & \textbf{0.936} & \textbf{0.695} & \textbf{0.908} & \textbf{0.770} & \textbf{0.692} & \textbf{0.948} & \textbf{0.796} & \textbf{0.916} \\
\bottomrule
\end{tabular}
}
\end{table*}

While Table~\ref{tab:main-results} contains the results of each active learning method with re-balanced mini-batches after bias correction, we also show the performance of each method without bias correction in Table~\ref{tab:main-results-no-bias-correction}. Recall that the purpose of the bias correction phase is to adjust the network outputs to the natural data distribution. On the full test set, our QBCat-Tail method performs worse than baseline active learning methods for the \gls{SPAA} and \gls{SPOP} question types. This is not surprising as our method prioritizes selecting tail data and without bias correction, it is not well-calibrated for the natural data distribution. When evaluated on the tail test set, QBCat-Tail outperforms all baselines. However, QBCat-Tail performs comparably to or outperforms other methods on both the full and tail test sets across question types after bias correction is applied in Table~\ref{tab:main-results} and Fig.~\ref{fig:additional-studies}.

\subsection{Active Learning Methods when Selecting Data from Only Tail Classes}

\begin{table*}[t]
\caption{$\Omega$ performance of each active learning method when methods \textbf{only select data from tail classes} over all increments when evaluated on the full and tail test sets. We report performance on each question type using both the AUROC and mAP metrics to compute $\Omega$. Each result is the average over 10 runs and computed based on 10 increments.
\vspace{0.1in}
\label{tab:main-results-tail-only}}
\centering
\resizebox{\linewidth}{!}{
\begin{tabular}{lcccccccccc}
\toprule
& \multicolumn{5}{c}{\textsc{\textbf{AUROC}}} & \multicolumn{5}{c}{\textsc{\textbf{mAP}}} \\
\cmidrule(r){2-6} \cmidrule(r){7-11}
\textsc{Model} & \gls{SPAS} & \gls{SPAA} & \gls{SPOS} & \gls{SPOP} & \gls{SPOO} & \gls{SPAS} & \gls{SPAA} & \gls{SPOS} & \gls{SPOP} & \gls{SPOO}  \\
\midrule
\multicolumn{2}{l}{\textit{Full Test Set}}\\
Random & 0.943 & \textbf{0.934} & \textbf{0.983} & \textbf{0.964} & \textbf{0.976} & 0.921 & 0.962 & \textbf{0.990} & 0.973 & \textbf{0.981} \\
Confidence & 0.944 & 0.933 & 0.981 & 0.963 & 0.974 & \textbf{0.922} & 0.962 & 0.986 & 0.972 & 0.979 \\
Entropy & \textbf{0.945} & \textbf{0.934} & 0.981 & \textbf{0.964} & 0.974 & \textbf{0.922} & \textbf{0.963} & 0.987 & \textbf{0.976} & 0.979 \\
Margin & 0.922 & 0.931 & 0.972 & 0.961 & 0.964 & 0.897 & \textbf{0.963} & 0.976 & 0.971 & 0.964 \\
QBCat-Tail & 0.941 & 0.927 & 0.978 & 0.951 & 0.974 & 0.919 & 0.959 & 0.985 & 0.958 & 0.979 \\
\midrule
\multicolumn{2}{l}{\textit{Tail Test Set}}\\
Random & 0.872 & 0.716 & \textbf{0.966} & \textbf{0.836} & \textbf{0.955} & 0.828 & \textbf{0.696} & \textbf{0.971} & \textbf{0.783} & \textbf{0.954} \\
Confidence & 0.876 & 0.712 & 0.963 & 0.834 & 0.951 & \textbf{0.830} & \textbf{0.696} & 0.966 & 0.782 & 0.949 \\
Entropy & \textbf{0.877} & \textbf{0.717} & 0.964 & \textbf{0.836} & 0.951 & \textbf{0.830} & \textbf{0.696} & 0.968 & \textbf{0.783} & 0.949 \\
Margin & 0.803 & 0.708 & 0.944 & 0.829 & 0.927 & 0.747 & \textbf{0.696} & 0.944 & 0.781 & 0.915 \\
QBCat-Tail & 0.866 & 0.695 & 0.959 & 0.804 & 0.951 & 0.819 & 0.695 & 0.965 & 0.776 & 0.949 \\
\bottomrule
\end{tabular}
}
\end{table*}

Since our QBCat-Tail active sampling method prioritizes data from tail classes during sampling, we were interested to see how other active learning methods performed when selecting data from only tail classes. In this experiment, we compared the Random, Confidence, Entropy, and Margin active sampling methods when selecting data from a pool consisting of unlabeled instances from only tail classes (i.e., we removed samples from head classes). Note that random sampling from tail data is equivalent to the QBCat-Tail (Freq.) method from Table~\ref{tab:main-results}. Results for each method using $\Omega$ when evaluated on the full and tail test sets are in Table~\ref{tab:main-results-tail-only}.

We find that all active sampling methods exhibit performance improvements when selecting samples from only tail data. This further supports the findings of our QBCat-Tail method. While performance is similar for several methods, QBCat-Tail does not require computing uncertainty scores for particular instances, making it simpler and more desirable to use in practice. 

It is worth noting that all methods exhibit improved performance on all question types on both test sets when selecting data from only tail classes (compared to selecting data from all classes in Table~\ref{tab:main-results}), further indicating the benefit of considering the class distribution during active learning on long-tailed datasets. While all methods exhibit improved performance when selecting data from tail classes, our QBCat-Tail approach is simplest since it does not require computing uncertainty scores at the instance level. This motivates the need for more methods that compute uncertainty scores at the class level instead of the instance level.
}

\end{document}